\pdfoutput=1

\documentclass[11pt]{article}

\usepackage[final]{acl}

\usepackage{times}
\usepackage{latexsym}

\usepackage[T1]{fontenc}

\usepackage[utf8]{inputenc}

\usepackage{microtype}

\usepackage{inconsolata}

\usepackage{hyperref}
\usepackage{url}

\usepackage{graphicx}
\usepackage{multirow}
\usepackage{amsmath}
\usepackage{amsthm}
\usepackage{algorithm}
\usepackage{algorithmic}
\usepackage{xcolor}  
\usepackage{amssymb}
\usepackage{amsfonts}
\usepackage{pifont}
\newcommand{\cmark}{\ding{51}}%
\newcommand{\xmark}{\ding{55}}%
\usepackage{pgfplots}
\usepackage{listings}
\definecolor{ace04}{HTML}{EC7829}
\definecolor{nyt}{HTML}{CB3236}
\definecolor{tgg}{HTML}{5B6AC0}
\definecolor{arg}{HTML}{9673A6}
\definecolor{16res}{HTML}{85B56A}

\usepackage{booktabs}
\title{Label Drop for Multi-Aspect Relation Modeling in Universal Information Extraction}

\author{Lu Yang$^{1}$, Jiajia Li$^{2,3}$, En Ci$^{1}$, Lefei Zhang$^{1}$, Zuchao Li$^{1,}$\thanks{$\ $  Corresponding author. This work was supported by the National Natural Science Foundation of China (No. 62306216, No. 72074171, No. 72374161), the Technology Innovation Program of Hubei Province (Grant No. 2024BAB043), the Natural Science Foundation of Hubei Province of China (No. 2023AFB816).}, Ping Wang$^{2,3}$\\
$^{1}$School of Computer Science, Wuhan University, Wuhan, China \\
$^{2}$School of Information Management, Wuhan University, Wuhan, China\\
$^{3}$Key Laboratory of Archival Intelligent Development and Service, NAAC \\
{\tt \{yang\_lu, cantata, cien.cs\}@whu.edu.cn} \\
{\tt \{zhanglefei, zcli-charlie, wangping\}@whu.edu.cn}
}

\begin{document}
\maketitle
\begin{abstract}
Universal Information Extraction (UIE) has garnered significant attention due to its ability to address model explosion problems effectively. 
Extractive UIE can achieve strong performance using a relatively small model, making it widely adopted. Extractive UIEs generally rely on task instructions for different tasks, including single-target instructions and multiple-target instructions. 
Single-target instruction UIE enables the extraction of only one type of relation at a time, limiting its ability to model correlations between relations and thus restricting its capability to extract complex relations.
While multiple-target instruction UIE allows for the extraction of multiple relations simultaneously, the inclusion of irrelevant relations introduces decision complexity and impacts extraction accuracy.
Therefore, for multi-relation extraction, we propose LDNet, which incorporates multi-aspect relation modeling and a label drop mechanism. By assigning different relations to different levels for understanding and decision-making, we reduce decision confusion. Additionally, the label drop mechanism effectively mitigates the impact of irrelevant relations. Experiments show that LDNet outperforms or achieves competitive performance with state-of-the-art systems on 9 tasks, 33 datasets, in both single-modal and multi-modal, few-shot and zero-shot settings.\footnote{https://github.com/Lu-Yang666/LDNet}
\end{abstract}

\section{Introduction}
Information Extraction (IE)~\citep{ie1,grishman_2019} tasks, both single-modal and multi-modal, encompass a wide variety of domains and relations between entities, leading to a highly diversified landscape. However, this diversification poses a significant challenge known as model explosion, which refers to the proliferation of models required to handle the diverse structures and relations present in different IE tasks. Traditionally, task-specific models~\citep{AdapCoAtt,j2,pure,ssegcn,peng2023novelenergybasedmodel, tian2023ngramunsupervisedcompoundationfeature,ziqi} have been developed to address the unique requirements of individual tasks. However, this approach is not scalable and becomes increasingly impractical as the number of tasks and their complexity grow. 


To tackle the issue of model explosion, Universal Information Extraction (UIE)~\citep{uie, knowcoder} has emerged as a promising paradigm. UIE aims to develop models that can extract information across different domains and relations, in both single-modal~\citep{lasuie} and multi-modal~\citep{tmr} setting, without relying on task-specific models for each individual task. By leveraging shared knowledge, UIE models can generalize well to various IE tasks, reducing the need for a multitude of specialized models.

Generative UIE~\citep{deepstruct, gollie} approaches have been explored, but their reliance on large generative models as the foundation limits their efficiency. These models suffer from computational complexity and resource requirements, hindering their practical applicability. In contrast, extractive UIE~\citep{uniex} approaches have gained popularity due to their ability to achieve strong performance using relatively small models.

Single-target instruction UIE~\citep{dygiepp} allows for the extraction of one type of relation at a time. While it excels in accuracy for simple relations, its limited efficiency and inability to model correlations between relations restrict its applicability to more complex IE tasks.
To address these limitations, multiple-target instruction UIE~\citep{mirror} has been proposed, enabling the extraction of multiple relations simultaneously. This approach aims to model correlations between relations and improve extraction efficiency. However, the incorporation of irrelevant relations introduces decision complexity and can have ramifications on extraction accuracy. 

Hence, to address the challenges in multi-relation extraction, we propose LDNet, a novel approach that leverages multi-aspect relation modeling and a label drop mechanism. In LDNet, we assign different relations to different levels for understanding and decision-making. This approach allows the model to capture the unique characteristics and nuances of each relation separately. By organizing relations into distinct levels, LDNet reduces decision confusion, enabling more accurate and reliable extraction results. Additionally, LDNet incorporates a label drop mechanism to address the impact of irrelevant relations. 
During the extraction process, LDNet selectively drops irrelevant labels, focusing on the most relevant relations for extraction. This mechanism helps mitigate the interference caused by irrelevant relations, ensuring that the model can concentrate its attention and resources on extracting the necessary and meaningful information. By filtering out noise and irrelevant signals, LDNet enhances the overall extraction performance and reduces the potential for false positives.

To assess the effectiveness of LDNet, we conduct extensive experiments on a diverse range of IE tasks and benchmark datasets, both single-modal and multi-modal. The evaluation covers few-shot and zero-shot settings to examine the generalization capability of LDNet. The results demonstrate that LDNet outperforms or achieves competitive performance compared to previous state-of-the-art systems across 9 tasks and 33 datasets.


\textbf{Our Contribution} 1) We propose LDNet, a novel approach that leverages multi-aspect relation modeling and a label drop mechanism. 2) We employ model transfer learning, a valuable strategy for further enhancing model performance across various datasets. 3) We conduct experiments on 33 datasets across 9 tasks, in both single-modal and multi-modal, few-shot and zero-shot settings, and the results demonstrate the superiority of LDNet.

\section{Related Work}\label{related-work}

\paragraph{Generative UIE} TANL~\citep{tanl} sees IE tasks as a sequence-to-sequence problem and utilizes T5 as the generative model. UIE~\citep{uie} also uses T5 as the backbone. In addition, UIE designs Structured Extraction Language (SEL) that can represent diversified IE tasks, thereby enabling it to perform on a wider range of IE tasks.
InstructUIE~\citep{instructuie} further incorporates the idea of instruction-tuning and utilizes FlanT5-11B~\citep{flan-t5} for IE tasks. DeepStruct~\citep{deepstruct} and GenIE~\citep{genie} both formulate the generated sequence as subject-relation-object triplets, with DeepStruct having a larger model size (10B). LasUIE~\citep{lasuie} proposes a novel structure-aware generative language model to unleash the power of syntactic knowledge. FSUIE~\citep{fsuie} introduces fuzzy span loss and fuzzy span attention to reduce over-reliance on span boundaries. 
GOLLIE~\citep{gollie} improves zero-shot results on unseen IE tasks by virtue of being fine-tuned to comply with annotation
guidelines.
TMR~\citep{tmr} addresses text-image misalignment by introducing a back-translation method using diffusion-based generative models. KnowCoder~\citep{knowcoder}  introduces a
code-style schema representation method. While the above generative UIE approaches offer a powerful solution for diversified IE tasks, they do not possess any notable advantages when it comes to efficiency. 
\paragraph{Extractive UIE} DyGIE++~\citep{dygiepp} utilizes a dynamic span graph to model long-range relations, and with graph propagation, the model can disambiguate challenging entity mentions.
UniEX~\citep{uniex} converts IE tasks into a token-pair problem, develops a traffine attention mechanism to integrate heterogeneous factors, and obtains the extraction target via a scoring matrix. These single-target extractive UIE approaches can achieve strong performance using a relatively small model; however, they lack the ability to model correlations between relations, thus limiting their capability to extract complex relations. 
\begin{figure*}[t]
    \centering
\includegraphics[width=0.95\textwidth]{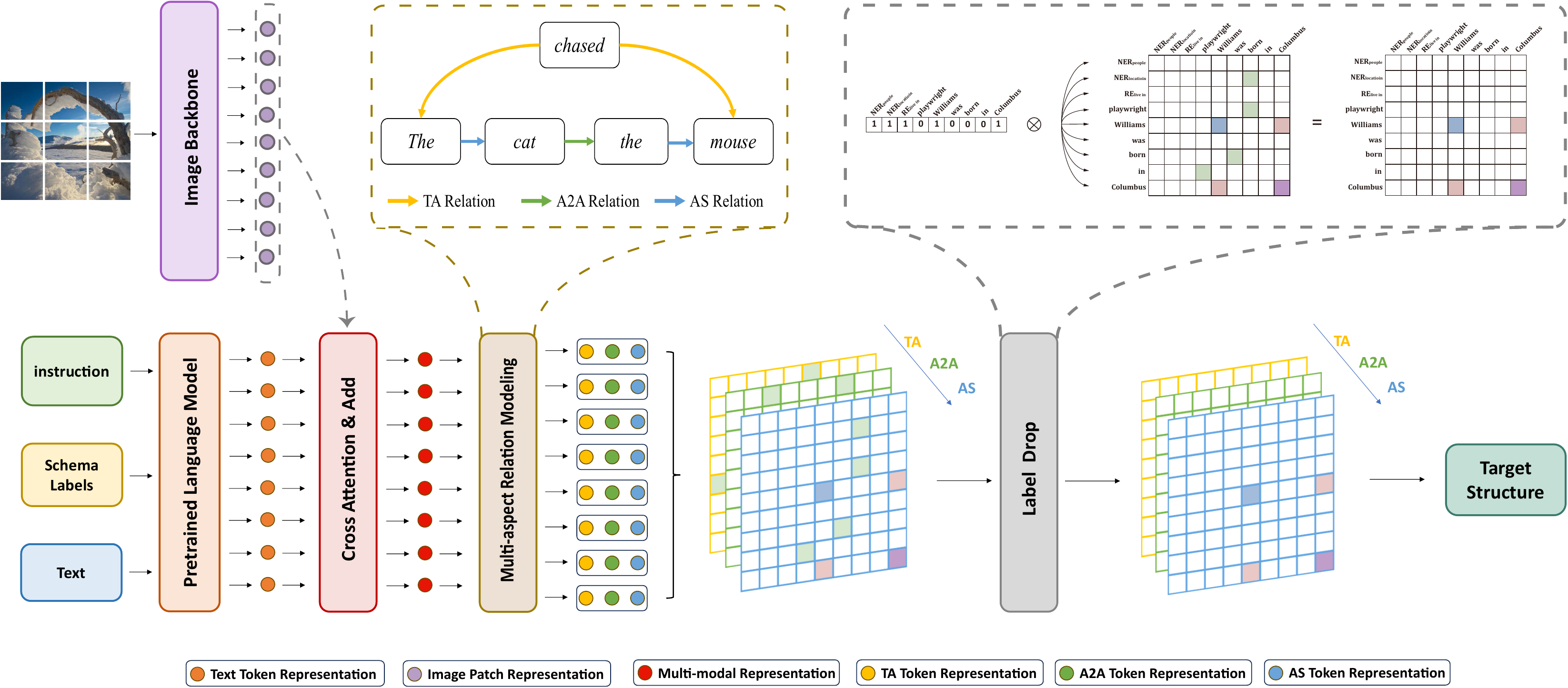}
    \caption{
         The overview framework of LDNet. LDNet constructs a unified input format, which combines instruction, schema labels, and text. The representation obtained from the PLM is fused with image representation obtained with the image backbone. The multi-modal representation is fed into the multi-aspect relation modeling component to produce probability matrices for TA, A2A, and AS relations, respectively. These matrices are then subjected to label drop to mask out non-existent relations. Finally, the probability matrices are fed into the decoding process to generate target structures.
    }
    \label{fig:model-framework}
\end{figure*}

OneIE~\citep{oneie} also uses a span graph, but unlike DyGIE++, it incorporates global features and adopts a CRF-based tagger to remove the constraint on the length of extracted mentions.
UMGF~\citep{umgf} adopts a unified intra-modal and inter-modal graph fusion method to represent visual and textual features within the same embedding space.
HVPNeT~\citep{hvpnet} designs pyramidal features for images, employing visual representations as insertable visual prefixes to guide error-insensitive predictive decisions of textual representations. 
MetaRetriever~\citep{metaretrieval} retrieves task-specific knowledge from pre-trained language models to enhance performance.
Mirror~\citep{mirror} transforms multiple tasks into a multi-span cyclic graph and predicts relations by verifying whether a cycle exists between slots in a tuple. While these multiple-target extractive UIE approaches take interactions between relations into account, they also include irrelevant relations, which leads to decision complexity and inaccuracy.

It is worth noting that the existing UIE models have basically only conducted experiments on single-modal or multi-modal IE tasks, and have not handled single-modal and multi-modal IE tasks simultaneously like LDNet.

\section{Methodology}\label{method-sec}


LDNet's overall framework is built upon a pretrained language model and an image backbone, consisting of a multi-aspect relation modeling component and a label drop mechanism, as shown in Figure~\ref{fig:model-framework}.

We formulate IE tasks as a multi-aspect span-based relation extraction problem. Specifically, we consider three kinds of relations among IE tasks: TA relation (trigger-to-argument relation), A2A relation (argument-to-argument relation), and AS relation (argument-span relation). The TA relation signifies the association of the trigger word with the identified span. The A2A relation describes the connection between two related spans, representing the semantic or contextual relation between the identified spans. The AS relation describes the connection within a span, enabling LDNet to analyze the internal structure and coherence within the span itself. 
As shown in Figure~\ref{fig:model-framework}, the AS relation is formed between ``the'' and ``cat'', the trigger word ``chased'' connects to ``the'' in the span ``the cat'' and ``mouse'' in the span ``the mouse'' through the TA relation, and ``cat'' is linked to ``the'' in the span ``the mouse'' through the A2A relation.

\subsection{Multi-aspect Relation Modeling}
\label{gp}

LDNet regularizes text input format into three components: instruction, schema labels, and text. Given an input sequence $\mathbf{x} = [x_1, x_2, \ldots, x_{|x|}]$, LDNet computes the text representation $\mathbf{H} = [h_1, h_2, \ldots, h_{|x|}]\in \mathbb{R}^{|x|\times d_h}$ as follows:
\begin{align}
\mathbf{H} &= PLM([x_1, x_2, \ldots, x_{|x|}]) \\
&= [h_1, h_2, \ldots, h_{|x|}]
\label{eq:1}
\end{align}
where $PLM(\cdot)$ is a pretrained language model.

To inject image information, 
given an image $I$, LDNet initially resizes it to $224\times224$, then divides it into $n_p$ patches according to the patch size specified by the image backbone, and subsequently derives image feature representation $\mathcal{V} \in \mathbb{R}^{n_p\times d_v}$ using the image backbone:
\begin{equation}
\mathcal{V} = VisionTransformer(I)
\label{eq:11}
\end{equation}
LDNet then employs cross-attention, where the image feature representation functions as the query and the text representation serves as both the key and value. The output of the attention mechanism is then subjected to the hyperbolic tangent activation function, followed by a summation operation. Finally, LDNet changes the sequence length of the resulting image feature representation, yielding the final image representation \(\mathbf{V} \in \mathbb{R}^{|x| \times d_h}\):
\begin{equation}
Q =\mathrm{FFNN_q^I}( MLP(\mathcal{V})),\; Q\in \mathbb{R}^{n_p \times d_h} 
\end{equation}
\begin{equation}
K=\mathrm{FFNN_k^I}(\mathbf{H}) ,\;
V=\mathrm{FFNN_v^I}(\mathbf{H}) 
\end{equation}
\begin{equation}
    \mathcal{V}'= \sum_{i=1}^{n_p} Tanh\left(Softmax\left(\frac{QK^{\mathrm{T}}}{\sqrt{d_h}}\right)V\right)
\end{equation}
\begin{equation}
\mathbf{V} = Tanh(\mathrm{FFNN_I}(\mathcal{V}'))
\label{eq:13}
\end{equation}
where $MLP\left ( \cdot  \right )$ represents a three-layer multilayer perceptron, $\mathrm{FFNN_{q/k/v}^I} \in \mathbb{R}^{d_h\times d_h}$ represents feed-forward network for generating query/key/value, and $\mathrm{FFNN_I}\in \mathbb{R}^{1\times |x|}$ represents the feed-forward network changing the sequence length of the resulting image feature representation.

LDNet modulates the fusion of image representation and text representation via a hyperparameter $\alpha \in [0,1]$, and the fused image-text representation $\mathbf{M}$ is expressed as:
\begin{equation}
\mathbf{M} = \mathbf{H} + \alpha \cdot \mathbf{V}
\label{eq:14}
\end{equation}
$\alpha$ is set to 0 when only doing single-modal IE tasks.

After obtaining the multi-modal representation $\mathbf{M}= [m_1, m_2, \ldots, m_{|x|}]$, LDNet utilizes Rotary Position Embedding (RoPE)~\citep{global_pointer} to achieve relative position encoding via combining the attention computation with absolute position encoding.
The queries and keys for different relations are calculated as follows:
\begin{equation}
q_i^r = \mathrm{FFNN}_q^r(m_i),\;
k_j^r = \mathrm{FFNN}_k^r(m_j)
\label{9}    
\end{equation}
where $r\in \left \{ TA, A2A, AS \right \} $, $\mathrm{FFNN}_{q/k}^r \in \mathbb{R}^{d_h\times d_i}$ are feed-forward layers for different relations, and $q_i^r$ and $k_j^r$ are the $i$-$th$ query and the $j$-$th$ key for different relations.

Afterwards, $q_i^r$ and $k_j^r$ are each left-multiplied by the transformation matrices $R_i$ and $R_j$ used in RoPE respectively. 
The dot product of $R_i$ and $R_j$ satisfies $R_i^{T} R_j = R_{j-i}$, thus incorporating relative position information. 
The probability $s_{ij}^r$ of the relation $r$ existing between the span from $i$ to $j$ is the scaled dot product of the transformed $q_i^r$ and $k_j^r$:


\begin{equation}
s_{ij}^r = \frac{(R_iq_i^r)^T(R_jk_j^r)}{\sqrt{d_i} }  
= \frac{q_i^{rT}R_{j-i}k_j^r}{\sqrt{d_i} } 
\label{eq:2}
\end{equation}

By parallelly computing the scaled dot product over all token pairs of different relations separately, we can obtain three probability matrices:
\begin{equation}
S^{r} = \begin{bmatrix}
 s_{11}^r &  s_{12}^r & \dots  &  s_{1|x|}^r\\
 s_{21}^r &  s_{22}^r& \dots  & s_{2|x|}^r\\
 \vdots   & \vdots  & \ddots  & \vdots \\
 s_{|x|1}^r & s_{|x|2}^r  & \dots & s_{|x||x|}^r
\end{bmatrix}^{} 
\label{eq:3}
\end{equation}
where $s_{ij}^r$ is the probability of the specific relation $r\in \left \{ TA, A2A, AS \right \} $ existing between the token pair $\left \langle x_{i},x_{j}  \right \rangle $.

During training, LDNet utilizes multi-label categorical cross-entropy loss as the loss function for multi-aspect relation modeling:
\begin{align}
l_{MR, neg}^r = log\left( 1+\sum_{\Omega _{neg}} e^{s_{ij}^r}\right) 
\end{align}
\begin{align}
    l_{MR, pos}^r = log\left ( 1+\sum_{\Omega _{pos}} e^{-s_{ij}^r}\right  )
    \end{align}
\begin{align}
    L_{MR} = \sum_{r \in \left \{ TA, A2A, AS \right \} }^{} \left(l_{MR, neg}^r + l_{MR, pos}^r \right) 
\end{align}
where $\Omega _{neg}$ and $\Omega _{pos}$ are the sets of negative and positive samples, respectively. Graph labels $G^r \in \mathbb{R}^{|x|\times |x|}$ are used to distinguish between negative and positive samples. Negative samples consist of position pairs where $G_{ij}^r = 0$, while positive samples are pairs where $G_{ij}^r = 1$. $G_{ij}^r$ represents the label of the token pair $\left \langle x_{i},x_{j}  \right \rangle $ for relation $r$.

\subsection{Label Drop}

To prioritize the relational token pairs, we employ label drop to filter out token pairs that are unlikely to have relations. 

Specifically, we first design a label vector $\mathbf{l}^r = [l_1^r, l_2^r, \ldots, l_{|x|}^r] \in \mathbb{R}^{1\times|x|}$ for each relation as the standard. We set the values of the elements in $\mathbf{l}^r$ whose indices fall within the gold spans and their corresponding schema labels to 1, and the rest to 0.

LDNet transforms the representation $\mathbf{M}$ into the predicted matrix $\hat{L}^r \in \mathbb{R}^{1\times|x|\times 1} $ via linear activation and Sigmoid function:
\begin{equation}
\hat{L}^r = Sigmoid\left ( \mathrm{FFNN}^r\left ( \mathbf{M} \right )  \right ) 
\label{eq:5}
\end{equation}
where $\mathrm{FFNN}^r \in \mathbb{R}^{d_h\times 1}$ is the feed-forward network for different relations. 

Later, LDNet squeezes the matrix $\hat{L}^r$ into $\hat{l}^r \in \mathbb{R}^{1\times|x|}$ and multiplies $\hat{l}^r$ with every row vector in the corresponding probability matrix $S^r$ computed in Section \ref{gp}, respectively, to obtain the final probability matrix $P^r$, which is later used for decoding:
\begin{equation}
p_{i\cdot }^r = \hat{l}^r \otimes s_{i\cdot }^r
\label{eq:7}
\end{equation}
where $p_{i\cdot }^r$ represents the $i$-$th$ row vector of $P^r$, $s_{i\cdot }^r$ represents the $i$-$th$ row vector of $S^r$, and $\otimes$ represents dot product. 
The detailed logic of label drop mechanism can be seen in Appendix~\ref{label-drop-logic}.


LDNet calculates the binary cross-entropy loss between $\hat{l}^r$ and $\mathbf{l}^r$ to make the predicted vector $\hat{l}^r$ approach the label vector $\mathbf{l}^r$ of the same relation during training:
\begin{equation}
l_{LD}^r =   -\frac{1}{n} \sum_{i}^{|x|} \left( \mathbf{l}_i^r\log(\hat{l}_i^r) + (1-\mathbf{l}_i^r) \log(1-\hat{l}_i^r) \right) \
\end{equation}
\begin{equation}
L_{LD} =\sum_{r \in \left \{ TA, A2A, AS \right \} }^{} l_{LD}^r 
\label{eq:6}
\end{equation}

After label drop, LDNet utilizes the three final probability matrices for relation decoding. 
During the process of relation decoding, LDNet extracts relation between token pair whose final probability is larger than the threshold of 0.5 and identifies potential relation structures. If LDNet detects a closed relation loop as shown in Figure~\ref{fig:model-framework}, it adds the extracted span to the predicted answer.

\subsection{Model Transfer Learning}

To further boost LDNet's performance, we propose a model transfer learning approach. We select the best-performance models fine-tuned on each dataset as the teacher models, and the generated probability distributions, namely $P^r$ of all data entries, from these teacher models are used as soft labels when fine-tuning the corresponding pre-trained student models.
Mean Squared Error (MSE) loss is employed to guide LDNet in reducing the discrepancy between the distributions of the student model and the teacher model:
\begin{equation}
L_{MT} =\sum_{r \in \left \{ TA, A2A, AS \right \} }^{}  \frac{1}{|x|^2} \sum_{i=1}^{|x|} \sum_{j=1}^{|x|} \left ( p_{ij}^r- \hat{p}_{ij}^r\right )^2 
\label{eq:8}
\end{equation}
where $p_{ij}^r$ represents the ${ij}$-${th}$ element of the $P^r$ generated by the teacher model, and $\hat{p}_{ij}^r$ represents the ${ij}$-${th}$ element of the $\hat{P}^r$ generated by the student model. The detailed algorithm is in Algorithm~\ref{alg:algorithm}.
\begin{algorithm}[t]
    \caption{Model Transfer Learning}
    \label{alg:algorithm}
    \textbf{Input}: Teacher model distributions $\mathcal{D_T}$ and pre-trained student model parameters $\theta$\\
    \textbf{Output}: Fine-tuned student model parameters $\Theta$\\
    \begin{algorithmic}[1]
  \FOR{$i$ in range (0, epochs)}
    \STATE (\textit{iterate over fine-tuning epochs})
    \FOR{$j$ in range (0, steps)}
          \STATE Obtain student model distributions $\mathcal{D_S}$.
  \STATE Set loss $L \gets L_{MR} + L_{LD}$.
    \FOR{$d_s \in \mathcal{D_S}$}
      \IF{$d_s$ finds the corresponding $d_t \in \mathcal{D_T}$}
        \STATE $L \gets L + MSE(d_s, d_t)$.
      \ENDIF
    \ENDFOR
        \STATE $\Theta = \theta - \gamma \frac{\hat{m_t}}{\sqrt{\hat{v_t}} + \epsilon}$  (use the AdamW optimizer to update parameters)
    \ENDFOR
  \ENDFOR
  \STATE \textbf{return} $\Theta$
\end{algorithmic}
\end{algorithm}
Thus, the complete objective for LDNet model training can be represented as follows:
\begin{equation}
    L = L_{MR} + L_{LD} + L_{MT}
\end{equation}

\begin{table*}[t]
  \centering
  \small
  \resizebox{\textwidth}{!}{%
  \begin{tabular}{c|c|ccccccccccc}
    \toprule
    Task & Datasets & TANL & UIE & DeepStruct & InstructUIE & USM &
      Mirror & FSUIE & UniEX & MetaRetriever & GoLLIE
     & \bf LDNet \\
    \midrule
    \multirow{3}{*}{NER}
     & ACE04   & -     & 86.89 & -     & - & 87.62 & 87.66 & 86.16 & 87.12 & 86.10 & - &\textbf{88.69} \\
     & ACE05   & 84.90 & 85.78 & 86.90 & 86.66 & 87.14 & 86.72& 86.91 & 87.02 & 84.01 & \textbf{88.10} &87.79 \\
     & CoNLL03 & 91.70 & 92.99 & 93.00 & 92.94 & 93.16 & 92.97 & - & 92.65& 92.38 & 92.80& \textbf{93.44} \\ 
      \midrule
      \multirow{4}{*}{RE}
     & ACE05   & 63.70 & 66.06 & 66.80 & - & 67.88 & 69.02& \textbf{74.16} & 66.06 &64.37& 63.60 &69.14 \\
     & CoNLL04 & 71.40 & 75.00 & 78.30 & 78.48 & 78.84 & 75.22 & -& 73.40 & 73.66& -  &\textbf{80.79} \\
     & NYT     & -     & 93.54 & 93.30 & 90.47 & 94.07 & 94.25 &- &- & -& - &\textbf{94.96} \\
     & SciERC  & -     & 36.53 & -     & 45.15 & 37.36 & 40.50 & - & 38.00 &35.77 & - &\textbf{46.85} \\
      \midrule
      \multirow{4}{*}{EE} &
      ACE05-Tgg  & 68.40 & 73.36 & 69.80 & 77.13 & 72.41 & 74.44 &- &74.08 &72.38 &72.20 & \textbf{83.56}\\
     & ACE05-Arg & 47.60 & 54.79 & 56.20 & \textbf{72.94} & 55.83 & 57.87 &- &53.92 & 52.62 & 66.00 &61.14\\
     & CASIE-Tgg & -     & 69.33 & -     & 67.80 & 71.73 & 73.09 &- &71.46 &69.76 &59.30 &\textbf{73.36} \\
     & CASIE-Arg & -     & 61.30 & -     & 63.53 & 63.26 & 61.27&- &62.91 & 60.37& 50.00&\textbf{63.88} \\
      \midrule
      \multirow{4}{*}{ABSA}
     & 14-res & - & 74.52 & - & - & 77.26 & 76.05& 74.17 & 74.77& 73.41 & - & \textbf{79.17} \\
     & 14-lap & - & 63.88 & - & - & 65.51 & 64.08  & 65.56 & 65.23 & 62.83 & - &\textbf{69.00} \\
     & 15-res & - & 67.15 & - & - & 69.86 & 67.41& 70.63 &68.58 & 65.85& - &\textbf{71.18} \\
     & 16-res & - & 75.07 & - & - & 78.25 &77.46 & 75.80 & 76.02 & 73.55 & - &\textbf{78.85}\\
     \midrule
     \multirow{3}{*}{MIE} & Twitter-2015 & - & - & - & - & - & 73.08* & -&- & -& - &\textbf{76.17}  \\
     & Twitter-2017 & - & - & - & - & - & 83.91* & -&- & -& - &\textbf{87.57}\\
     & MNRE & - & - & - & - & - & 70.58* & -&- & -& - &\textbf{75.79}\\
    \bottomrule
    \end{tabular}%
  }
  \caption{Results on 16 IE benchmarks. -Tgg and -Arg refer to trigger F1 score and argument F1 score, respectively. Mirror does not test on multi-modal IE datasets. The results marked with * are the performance we obtain using Mirror's model and training checkpoint.}
  \label{tab:main-results}
\end{table*}

\begin{table*}[t]
  \centering
  \resizebox{\textwidth}{!}{%
  \begin{tabular}{c|c|cccc|cccc|cc}
    \toprule
    \multirow{2}{*}{Task} & \multirow{2}{*}{Datasets} & Mirror &
      Mirror &
      Mirror &
      Mirror &
      $\mathrm{LDNet}_{MT^-}$ &
      $\mathrm{LDNet}_{MT^-}$ &
      $\mathrm{LDNet}_{MT^-}$ &
      $\mathrm{LDNet}_{MT^-}$ &
      LDNet &
      LDNet\\
    & & \shortstack{w/ PT \\ w/ Inst.} & \shortstack{w/ PT \\ w/o Inst.} & \shortstack{w/o PT \\ w/ Inst.} & \shortstack{w/o PT \\ w/o Inst.} & \shortstack{w/ PT \\ w/ Inst.} & \shortstack{w/ PT \\ w/o Inst.} & \shortstack{w/o PT \\ w/ Inst.} & \shortstack{w/o PT \\ w/o Inst.} & \shortstack{w/ PT \\ w/ Inst.} & \shortstack{w/ PT \\ w/o Inst.}\\
    \midrule
    \multirow{3}{*}{NER}
     & ACE04   &  87.16 & 86.39 & 87.66 & 87.26 & 85.68 & 85.63 &88.21 & 86.76 & 86.94 &\textbf{88.69}\\
     & ACE05   & 85.34 & 85.70 & 86.72 & 86.45 & 84.70 & 85.87 & 87.11 & 85.38 &87.70 &\textbf{87.79} \\
     & CoNLL03 & 92.73 & 91.93 & 92.11 & 92.97 & 92.23 & 92.67 & 93.30 & 92.70 & \textbf{93.44} & 92.81\\ 
      \midrule
      \multirow{4}{*}{RE}
     & ACE05   & 67.86 & 67.86 & 64.88 & 69.02 & 66.35 & 67.73 & 69.10 & 68.58 & 69.02 &\textbf{69.14} \\
     & CoNLL04 & 75.22 & 72.96 & 71.19 & 73.58 & 76.90 & 78.26 & 75.81 & 75.03 & \textbf{80.79} & 80.37\\
     & NYT     & 93.85 & 94.25 & 93.95 & 93.31 & 94.00 & 94.34 & 94.13& 93.73 & 92.46 & \textbf{94.96} \\
     &SciERC  & 36.89 & 37.12 & 36.66 & 40.50 & 43.79 & 45.58 & 43.48 & 46.06 & 42.79 & \textbf{46.85}\\
      \midrule
      \multirow{4}{*}{EE} &
      ACE05-Tgg  & 74.44 & 73.05 & 72.66 & 73.38 &73.99 &75.81 & 72.92& 73.98 & 72.34 & \textbf{83.56}  \\
     & ACE05-Arg & 55.88 & 54.73 & 56.51 & 57.87 &56.27 &57.88 &50.70 & 58.01 & 54.80 & \textbf{61.14}\\
     & CASIE-Tgg & 71.81 & 71.60 & 73.09 & 71.40 & 70.87 & 71.23 & 71.97 & 73.18 & 71.86 & \textbf{73.36}\\
     & CASIE-Arg & 61.27 & 61.04 & 60.44 & 58.87 & 61.34 & 61.92 & 62.33 & 63.20 & 61.51 & \textbf{63.88}\\
      \midrule
      \multirow{4}{*}{ABSA}
     & 14-res & 75.06 & 74.24 & 76.05 & 75.89 & 73.93 & 74.76 & 76.16 &77.93 &76.40 & \textbf{79.17} \\
     & 14-lap & 64.08 & 62.48 & 59.56 & 60.42 & 66.60 & 66.03 &65.32 &63.80 & \textbf{69.00} &65.42 \\
     & 15-res & 66.40 & 63.61 & 60.26 & 67.41 & 66.63 & 66.87 & 66.30 &65.79 &69.51 & \textbf{71.18}\\
     & 16-res & 74.24 & 75.40 & 73.13 & 77.46 & 74.41 & 76.10 & 77.97 &77.19 & 76.34 & \textbf{78.85}\\
     \midrule
     \multicolumn{2}{c|}{Avg.} & 72.15 & 71.49 & 70.99 & 72.39 &72.51 & 73.38 & 72.99 & 73.42 & 73.66 & 75.81\\
    \bottomrule
    \end{tabular}%
  }
  \caption{Results of LDNet compared with Mirror. PT stands for pre-training, and Inst. represents the task instruction. $\mathrm{LDNet}_{MT^-}$ denotes LDNet without model transfer learning.}
  \label{tab:compare-mirror}
\end{table*}%
\begin{table}[]
  \centering
\resizebox{0.7\columnwidth}{!}{%
  \small
  \begin{tabular}{lrrr}
    \toprule
    & \multicolumn{1}{c}{P} & \multicolumn{1}{c}{R} & \multicolumn{1}{c}{F1} \\
    \midrule
    \multicolumn{4}{l}{\textit{Discontinuous NER: CADEC}} \\
    BART-NER & 70.08 & 71.21 & 70.64 \\
    W2NER    & 74.09 & 72.35& 73.21 \\
    Mirror$_{\text{w/ PT \& Inst.}}$   & 74.83 & 65.45 & 69.83 \\
    Mirror$_{\text{w/o PT \& Inst.}}$ & 68.80 & 68.38 & 68.59 \\
    LDNet$_{\text{w/ PT \& Inst.}}$ & \textbf{84.82} & 71.16 & \textbf{77.39}\\
    LDNet$_{\text{w/o PT \& Inst.}}$& 71.89 & 68.18  & 69.98  \\
    \midrule
    \multicolumn{4}{l}{\textit{N-ary Tuples: HyperRED}} \\
    CubeRE   & 66.39 & 67.12 & 66.75 \\
    Mirror$_{\text{w/ PT \& Inst.}}$   & 71.29 & 62.46 & 66.58  \\
    Mirror$_{\text{w/o PT \& Inst.}}$ & 75.41 & 61.14 & 67.53  \\
    LDNet$_{\text{w/ PT \& Inst.}}$ & 69.39 & 66.56 & \textbf{67.95}\\
    LDNet$_{\text{w/o PT \& Inst.}}$& 68.40 & 64.31 & 66.29 \\
    \bottomrule
  \end{tabular}%
  }
  \caption{
    Results on multi-span and n-ary information extraction tasks.
  }
  \label{tab:multi-span-n-ary-results}    
\end{table}
 \begin{table*}[]
    \centering
    \resizebox{0.75\textwidth}{!}{%
    \centering
    \small
    \begin{tabular}{l|c|cccccccc}
        \toprule
    Model & Parameter Scale &
      \multicolumn{1}{c}{Movie} & 
      \multicolumn{1}{c}{Restaurant} &
      \multicolumn{1}{c}{AI} &
      \multicolumn{1}{c}{Literature} &
      \multicolumn{1}{c}{Music} &
      \multicolumn{1}{c}{Politics} &
      \multicolumn{1}{c}{Science} &
      \multicolumn{1}{c}{Avg.} \\
      \midrule
    USM          & 372M   & 37.73          & 14.73          & 28.18          & \bf 56.00 & 44.93          & 36.10          & 44.09          & 37.39          \\
    InstructUIE    & 11B  & \textbf{63.00} &  20.99    & 49.00          & 47.21          & 53.61          & 48.15          & 49.30          & 47.32         \\
    Mirror & 304M & 39.20 & 16.32 & 45.23 & 46.32 & 58.61 & 67.30 & 54.84 & 46.83  \\
    LDNet & 304M & 41.92 & \textbf{22.91} & \textbf{49.02} & 55.11 & \textbf{61.10} & \bf 69.03 & \textbf{59.83}& \textbf{51.27} \\
     \midrule
      Llama-3 & 8B & 7.48* & 6.15* & 7.40* & 5.81* & 3.41* & 8.55* & 4.43* & 6.18*\\
    \bottomrule
    \end{tabular}%
    }
    \caption{
        Zero-shot results on 7 NER datasets. The results for Llama-3 are obtained from our experiments and are for reference only.}
    \label{tab:zero_shot_ner}
\end{table*}
\begin{table}[t]
    \centering
    \resizebox{0.85\columnwidth}{!}{%
    \begin{tabular}{c|c|ccc|c}
    \toprule
    Task & Model  & 1-shot & 5-shot & 10-shot & Avg.  \\
    \midrule
    \multirow{4}{*}{\shortstack{NER\\CoNLL03}}
     & UIE    & 57.53  & 75.32  & 79.12   & 70.66 \\
     & USM    & 71.11  & 83.25  & 84.58   & 79.65 \\
     & Mirror & 76.49 & 82.45 & 84.69 & 81.21 \\
     & LDNet & \bf 78.33  & \bf 84.53 & \bf 85.39& \bf 82.75\\
    \midrule
    \multirow{4}{*}{\shortstack{RE\\CoNLL04}}
    & UIE    & 34.88  & 51.64  & 58.98   & 48.50 \\
    & USM    & 36.17  & 53.20  & 60.99   & 50.12 \\
    & Mirror &     26.29   &   47.42     &   55.77      &  43.16     \\
    & LDNet & \bf 37.93 & \bf 53.74 & \bf 61.46 &\bf 51.04\\
    \midrule
    \multirow{4}{*}{\shortstack{Event Trigger\\ACE05}}
    & UIE    & 42.37  & 53.07  & 54.35   & 49.93 \\
    & USM    & 40.86  & 55.61  & 58.79   & 51.75 \\
    & Mirror &   47.77    & 57.90       &  59.16       & 54.94      \\
    &LDNet & \bf 54.77 & \bf 62.75 & \bf 64.24 & \bf 60.59\\
    \midrule
    \multirow{4}{*}{\shortstack{Event Arg\\ACE05}}
    & UIE    & 14.56  & 31.20  & 35.19   & 26.98 \\
    & USM    & 19.01  & 36.69  & 42.48   & 32.73 \\
    & Mirror &    23.18    & 37.74       & 39.20        & 33.38     \\
    &LDNet &\bf 25.42 & \bf 39.12 & \bf 43.04 & \bf 35.86\\
    \midrule
    \multirow{4}{*}{\shortstack{ABSA\\16-res}}
    & UIE    & 23.04  & 42.67  & 53.28   & 39.66 \\
    & USM    & 30.81  & 52.06  & 58.29   & 47.05 \\
    & Mirror &      36.21  & 51.65       & 58.59        & 48.82 \\
    & LDNet & \bf 40.43 &\bf  55.29& \bf 60.20 &  \bf 51.97 \\
    \bottomrule
    \end{tabular}%
    }
    \caption{
        Few-shot results on 4 IE tasks. These datasets are not included in pre-training, and LDNet does not apply model transfer learning in this setting, thus avoiding the risk of information leakage. 
    }
    \label{tab:few_shot}
    \end{table}
    
\section{Experiments}\label{experiments}

\subsection{Experiment Setup}
We use DeBERTa-v3-large~\citep{deberta} as the PLM, ViT~\citep{vit} as the image backbone, and AdamW~\citep{adamw} as the optimizer. We conduct experiments on ACE04~\citep{ace04}, ACE05~\citep{ace05}, CoNLL03~\citep{conll03}, CoNLL04~\citep{conll04}, NYT~\citep{nyt}, SciERC~\citep{scierc}, CASIE~\citep{casie}, 14-res and 14-lap~\citep{absa14}, 15-res~\citep{absa15}, 16-res~\citep{absa16}, Twitter2015~\citep{twitter15}, Twitter2017~\citep{twitter17}, and MNRE~\citep{mnre} datasets. We used the F1 score as the metric unless otherwise specified. 

We compare LDNet with TANL~\citep{tanl}, DeepStruct~\citep{deepstruct}, UIE~\citep{uie}, InstructUIE~\citep{instructuie}, USM~\citep{usm}, Mirror~\citep{mirror}, FSUIE~\citep{fsuie}, UniEX~\citep{uniex}, MetaRetriever~\citep{metaretrieval}, and GoLLIE~\citep{gollie} in single-modal IE tasks, and with Mirror in Multi-modal Information Extraction (MIE) tasks. 

In different pre-training and fine-tuning strategies, we specifically compare LDNet with Mirror. The pre-training of LDNet is before fine-tuning on downstream datasets, the pre-training datasets and hyperparameters are in Appendix~\ref{hyperparams}. 
`With and without pre-training (PT)' refers to whether LDNet underwent pre-training on a pre-training dataset before fine-tuning on the downstream dataset. `With and without task instructions (Inst.)' indicates whether the instruction part of LDNet’s input is an empty string. In the configuration `w/ PT w/o Inst.', the instruction part is an empty string, while in `w/ PT w/ Inst.', the instruction part is not empty and resembles a string like `Please determine the two entities mentioned in the text and specify the nature of their relationship.'
The best results we obtained in the main results of Table~\ref{tab:main-results} are from the four configurations: `w/ PT w/ Inst.', `w/ PT w/o Inst.', `w/o PT w/ Inst.', and `w/o PT w/o Inst.'

BART-NER~\citep{bart}, W2NER~\citep{w2ner}, and Mirror are chosen as the baseline models for the multi-span discontinuous NER task. In the hyper RE task, we select CubeRE~\citep{hyperred} and Mirror as the baseline models. As for the MRC tasks, We compare LDNet to BERT~\citep{bert}, RoBERTa~\citep{roberta},  DeBERTa-v3~\citep{deberta}, and Mirror.

\subsection{Main Results}

LDNet main results over 16 IE and MIE datasets are shown in Table~\ref{tab:main-results} and Table~\ref{tab:compare-mirror}. We can observe that: 

1) By adopting multi-aspect relation modeling and applying label drop on separate probability matrix, LDNet offers an effective methodology for IE and MIE. LDNet achieves state-of-the-art performance across almost all datasets and tasks. Although LDNet slightly underperforms FSUIE on ACE05-RE and underperforms GoLLIE on ACE05-NER, it surpasses them in other tasks and covers a broader range of tasks, including MRC, classification, discontinuous NER, and hyper-RE.
MRC and multi-modal results are in Appendix~\ref{multi-modal baselines}.

2) Despite having a relatively small model scale, LDNet consistently delivers superior results across almost all IE tasks. LDNet outperforms DeepStruct (10B) in all tasks.
The comparison of pre-trained language model parameter scales is included in Appendix~\ref{model-parameters-and-computational-complexity} and more comparisons with models of different scales can be found in the Appendix~\ref{multi-modal baselines}.
We also conduct ablation studies on LDNet, focusing on pre-training and fine-tuning strategies, as shown in Table~\ref{tab:compare-mirror}. LDNet surpasses Mirror in almost all settings and across all datasets. 

3) Model transfer learning provides a valuable strategy for enhancing performance across all datasets, enabling LDNet to leverage information sharing between teacher models and student models.

\begin{figure*}
\begin{minipage}{0.38\textwidth}
\centering
\resizebox{\textwidth}{!}{
\begin{tabular}{cl|cc|cc}
\toprule
 & & \multicolumn{2}{c|}{\text{w/ PT \& Inst.}} &  \multicolumn{2}{c}{\text{w/ PT \& w/o Inst.}} \\ 
\multicolumn{2}{c|}{w/ Label Drop} & \textcolor{red}{\xmark}  & \color[HTML]{008114}\cmark  & \textcolor{red}{\xmark}  & \color[HTML]{008114}\cmark \\ 
\midrule
\multicolumn{1}{c|}{\multirow{2}{*}{NER}}   & ACE05      & 87.12 & \textbf{87.20} & 86.45 & \textbf{86.97}     \\
 \multicolumn{1}{l|}{}                    & CoNLL03    & 93.75 & \textbf{94.60} & 94.93 &\textbf{95.11}\\ 
 \midrule
\multicolumn{1}{c|}{\multirow{2}{*}{RE}}    & NYT        &93.09& \textbf{94.43} & 93.94 & \textbf{94.41}        \\ 
 \multicolumn{1}{l|}{}                      & SciERC     & 43.75 & \textbf{43.94} & 38.71 & \textbf{46.27}     \\ 
 \midrule
\multicolumn{1}{c|}{\multirow{2}{*}{EE}}   & ACE05-Tgg  & 63.55 & \textbf{78.74} & 70.40 & \textbf{80.23}     \\
\multicolumn{1}{l|}{}                       & ACE05-Arg  & 49.37 & \textbf{55.78} & 46.67 & \textbf{62.76}        \\
 \midrule
\multicolumn{1}{l|}{\multirow{4}{*}{ABSA}} & 14-res     & 82.18 & \textbf{83.58} & 83.70 & \textbf{84.79}\\
\multicolumn{1}{l|}{}                      & 14-lap     & 77.23&\textbf{78.11} & 71.12 & \textbf{71.72} \\
\multicolumn{1}{l|}{}                      & 15-res     & 63.91 & \textbf{67.38} & 65.23 & \textbf{70.36} \\
\multicolumn{1}{l|}{}                      & 16-res     & 76.25 & \textbf{78.15} & 77.24 & \textbf{78.17} \\ 
\bottomrule
\end{tabular} 
}
    \label{tab:label_drop}
\end{minipage}
\begin{minipage}{0.30\textwidth}
  \includegraphics[width=\textwidth]{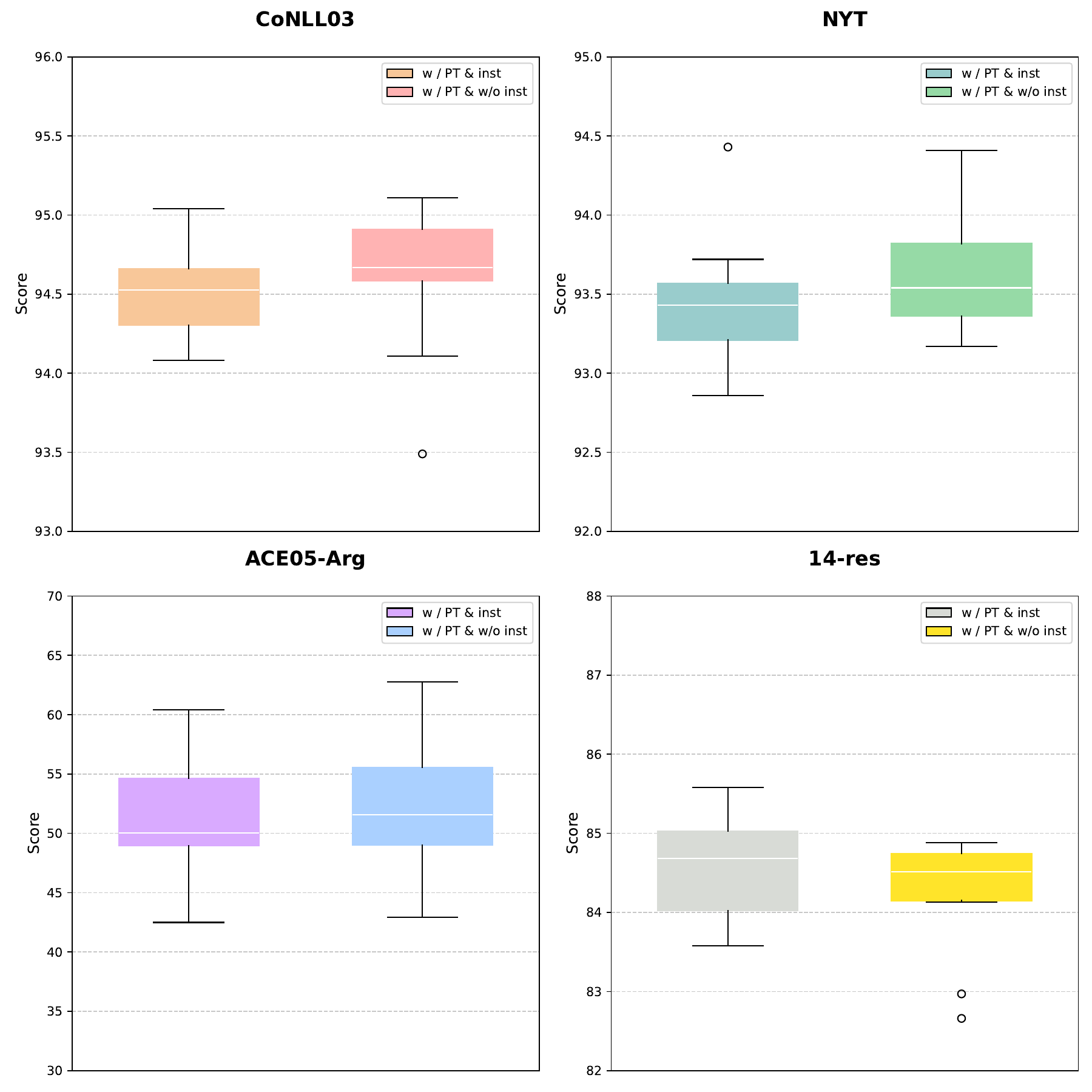}
\end{minipage}
\begin{minipage}{0.31\textwidth}
  \includegraphics[width=\textwidth]{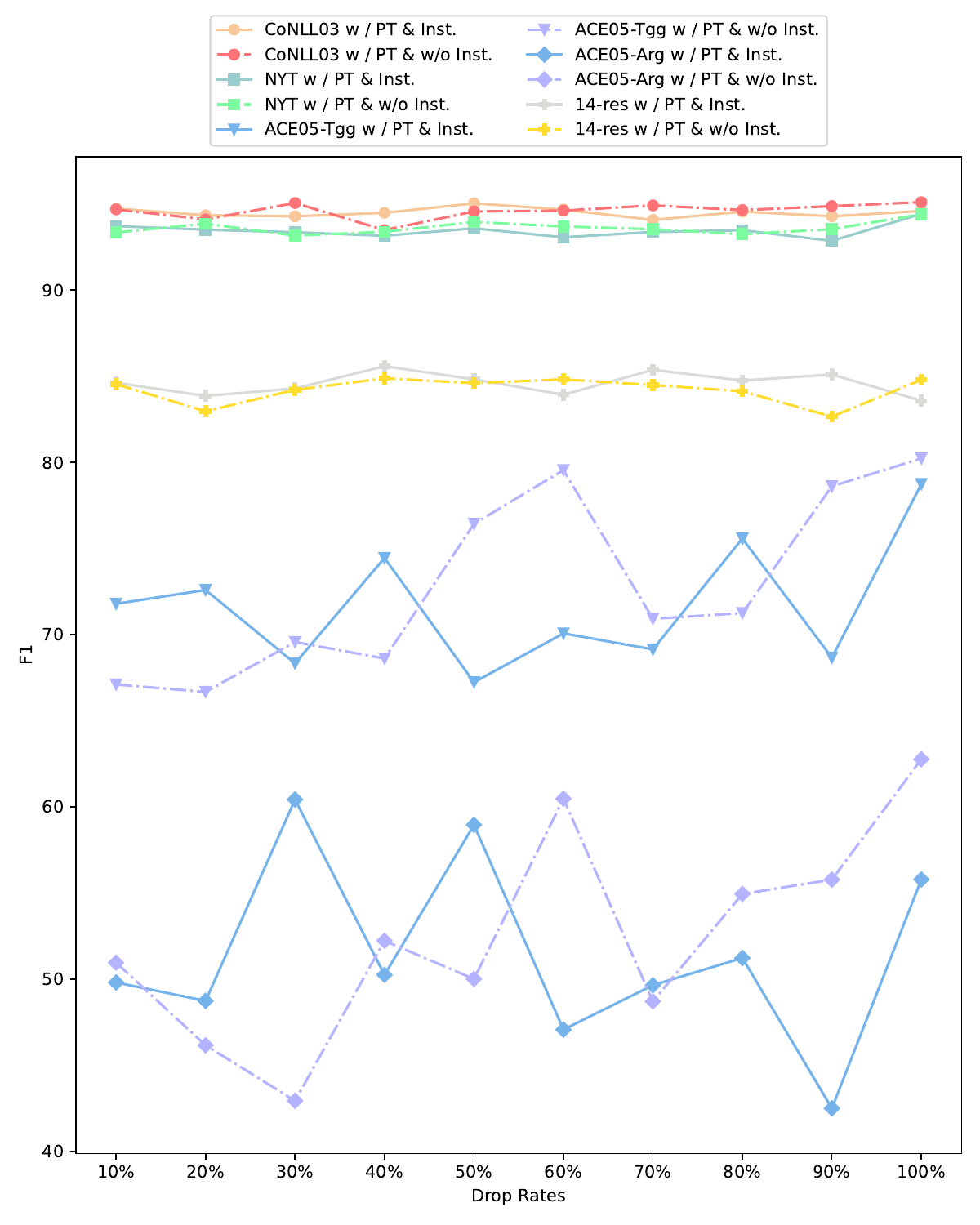}
\end{minipage}
    \caption{
        Results of the ablation study on the label drop mechanism. 
        Table on the left shows LDNet's performance with and without the Label Drop mechanism, box plot in the middle and line chart on the right illustrate LDNet's performance under different drop rates.
    }
    \label{fig:image_and_table}
\end{figure*}

Besides the triplet-based IE and MIE tasks, LDNet also demonstrates its effectiveness in discontinuous NER and n-ary hyper RE tasks. We provide results with pre-training (w/ PT, w/ Inst) and without pre-training (w/o PT, w/ Inst). As presented in Table~\ref{tab:multi-span-n-ary-results}, LDNet achieves improvements over previous methods, increasing by 4.18\% and 0.42\% on the CADEC~\citep{cadec} and HyperRED~\citep{hyperred} datasets, respectively.



\subsection{Few-shot Results \& Zero-shot Results }

Followed by \citet{mirror}, we analyze LDNet's quick adaptation ability on NER, RE, EE, and ABSA tasks under 1-shot, 5-shot, and 10-shot settings.
We compared LDNet with strong baselines such as UIE, USM, and Mirror. Table~\ref{tab:few_shot} shows the superior performance of LDNet under low-resource settings. On average, LDNet improves results by 3.41\%, 2.26\%, and 1.68\% under 1-shot, 5-shot, and 10-shot settings, respectively. Additionally, LDNet achieves average improvements of 1.27\%, 0.92\%, 3.52\%, and 3.02\% over the NER, RE, EE, and ABSA tasks, respectively.



We also examine LDNet's extendability in the zero-shot setting using NER datasets from 7 distinct fields~\citep{mit_ner_corpus,crossner}. LDNet is compared with USM, InstructUIE, Mirror and Llama-3-8B~\citep{llama3}. 
The zero-shot results are presented in Table~\ref{tab:zero_shot_ner}.
LDNet consistently outperforms Mirror and Llama-3-8B across all evaluated datasets. Although LDNet scores 0.89 lower than USM on the Literature dataset and is slightly lower than InstructUIE on the Movie dataset, the parameter scale of LDNet's pre-trained language model is smaller than that of USM's, and much smaller than that of InstructUIE's. Additionally, LDNet still achieves the highest average performance overall.

\subsection{Analysis on Label Drop}
To investigate the effectiveness of our label drop mechanism, we conduct ablation studies under two settings: with and without label drop. 
We fine-tune for 100 epochs on the IE datasets in both of the settings to fully exploit the potential of the label drop mechanism.
And in order to better demonstrate the effectiveness of the label drop mechanism, we do not perform model transfer learning in the ablation study.
The results are shown in Figure~\ref{fig:image_and_table}. Compared to the results of LDNet with only the multi-aspect relation modeling component, the model with the label drop mechanism shows improved performance on most datasets. It achieves an improvement of 1.93\% on average for the ABSA task and a substantial increase of 11.88\% on the ACE05 dataset for the EE task.

To further analyze the capability of the label drop mechanism, we conduct experiments on 4 text datasets involving different tasks. We randomly drop different portions of the probability matrix and test the performance. We test drop rates of 10\%, 20\%, 30\%, 40\%, 50\%, 60\%, 70\%, 80\%, 90\%, and 100\%. We also fine-tune 100 epochs for the analysis. The results are shown in Figure~\ref{fig:image_and_table}. 
More detailed results can be found in the Appendix~\ref{drop-rate ablation}. 

We can see that LDNet exhibits stable and strong performance on the NER, RE, and ABSA tasks, under different drop rates, demonstrating the robustness of the label drop mechanism. For the EE task, the performance is more volatile, with the best performance occurring under the full drop setting, which is as expected.
The fluctuations in performance on the ACE05 dataset in Figure~\ref{fig:image_and_table} are due to the large variety of label texts in the schema labels. The label texts following $\texttt{[LM]}$ include 33 types, such as `transport', `elect', `start position', `nominate', `end position', `attack', `meet', `marry', `phone write', and so on. The label texts following $\texttt{[LR]}$ have even more varieties, totaling 104 types. When only part of the probability matrix is dropped, some irrelevant token pairs' TA, A2A, or AS relation probabilities may not be dropped and could still exceed the threshold. As a result, non-existent TA, A2A, and AS relations may be included in the decoding process, and if they form a cycle, non-existent relations are predicted. 
Additionally, since we are \textbf{randomly} dropping portions  of the probability matrix, even with a low drop rate, there is still a chance to accurately filter out token pairs that are unlikely to have relations. The content of the schema labels can be found in Appendix~\ref{appendix-schema-labels}, and the specific process of relation cycling can be referenced in Appendix~\ref{appendix-unknown-schemas} or Appendix~\ref{appendix-discontinuousNER}. The stable high performance of other datasets is attributed to the significantly fewer types of label texts; for example, the CoNLL03 dataset has only four types: `miscellaneous', `person', `location', and `organization'. Even with a very low drop rate like 10\%, the difficulty of filtering out non-existent relations is much lower in such a small range.

\begin{table}[t]
  \centering
  \resizebox{0.7\columnwidth}{!}{%
  \begin{tabular}{c|c|c}
    \toprule
    Task & Datasets & Label Drop Accuracy\\
    \midrule
    \multirow{1}{*}{NER}
     & CoNLL03 & 93.04 \\ 
      \midrule
      \multirow{2}{*}{RE}
     & ACE05   & 94.69 \\
     & CoNLL04 & 87.98 \\
      \midrule
      \multirow{4}{*}{ABSA}
     & 14-lap & 92.61 \\
     & 15-res & 89.82 \\
     & 16-res & 90.41 \\
     \midrule
     \multicolumn{2}{c|}{Avg.} & 91.76 \\
    \bottomrule
    \end{tabular}%
  }
  \caption{Results of the accuracy of label drop mechanism. }
  \label{label-drop-accuracy}
\end{table}%
\begin{table}[t]
  \centering
  \resizebox{\columnwidth}{!}{%
  \begin{tabular}{c|c|c|c|c|c}
    \toprule
    \multirow{2}{*}{Task} & \multirow{2}{*}{Datasets} &
      Mirror & $\mathrm{LDNet}_{MR}$ & $\mathrm{LDNet}_{LD}$ & 
      $\mathrm{LDNet}_{MT^-}$  \\
    & & \shortstack{w/o PT \\ w/ Inst.} & \shortstack{w/o PT \\ w/ Inst.} & \shortstack{w/o PT \\ w/ Inst.} & \shortstack{w/o PT \\ w/ Inst.}  \\
    \midrule
    \multirow{1}{*}{NER}
     & CoNLL03 &  92.11 & 92.84 & 93.24 & \textbf{93.30} \\ 
      \midrule
      \multirow{2}{*}{RE}
     & ACE05   & 64.88 & 65.76 & 68.09 & \textbf{69.10} \\
     & CoNLL04 & 71.19 & 75.11 &  71.97 &  \textbf{75.81}  \\
      \midrule
      \multirow{4}{*}{ABSA}
     & 14-lap & 59.56 & 62.94 & 60.75 & \textbf{65.32} \\
     & 15-res & 60.26 & 61.25 & 60.69 & \textbf{66.30} \\
     & 16-res & 73.13 & 76.81 & 75.76 & \textbf{77.97} \\
     \midrule
     \multicolumn{2}{c|}{Avg.} & 70.19 & 72.12 & 71.75 & \textbf{74.30} \\
    \bottomrule
    \end{tabular}%
  }
  \caption{Results of the ablation study on the multi-aspect relation modeling mechanism. $\mathrm{LDNet}_{MR}$ represents LDNet using only Multi-aspect Relation Modeling, $\mathrm{LDNet}_{LD}$ represents LDNet using only Label Drop, and $\mathrm{LDNet}_{MT^-}$ indicates LDNet using both Multi-aspect Relation Modeling and Label Drop. }
  \label{multi-aspect modeling ablation}
\end{table}%
We also evaluate the accuracy of the label drop mechanism separately. The label drop probabilities $\hat{l}^r$ can be used to assess the accuracy of the label drop model. The accuracy can be computed as follows: we repeat $\hat{l}^r \in \mathbb{R}^{1\times|x|}$ to match the shape of the label matrices to create probability matrices $A^r \in \mathbb{R}^{|x|\times |x|}$, $r \in \left \{ TA, A2A, AS \right \}$. Values in $A^r$ below 0.5 are considered 0, while values above 0.5 are considered 1. We then compare this to the label matrices, and the number of correct values divided by the total number of values gives the accuracy. We test the accuracy, and the results are shown in the Table~\ref{label-drop-accuracy}. It can be seen that the label drop accuracy is generally high, with an average accuracy exceeding 90\%.

\subsection{Ablation Results of Multi-aspect Relation Modeling} We further conduct experiments on the Multi-aspect Relation Modeling mechanism. In these experiments, we utilize only the Multi-aspect Relation Modeling mechanism of the LDNet model, without incorporating Label Drop or Model Transfer Learning. We report the performance of LDNet under the `w/o PT w/ Inst.' setting, with the selected comparison baseline being the performance of Mirror under the same setting. 
The results are shown in the Table~\ref{multi-aspect modeling ablation}. As can be seen, $\mathrm{LDNet}_{MR}$ still outperforms Mirror under the same setting, demonstrating the effectiveness of the Multi-aspect Relation Modeling mechanism. The presence of Multi-aspect Relation Modeling reduces decision confusion while also creating a more suitable environment for the label drop mechanism. So it can be shown in the table that $\mathrm{LDNet}_{MT^-}$ achieves the best performance among the three: $\mathrm{LDNet}_{MR}$, $\mathrm{LDNet}_{LD}$, and $\mathrm{LDNet}_{MT^-}$.

\section{Conclusion and Discussion}\label{conclusion}
In this paper, we propose LDNet, a novel network that combines multi-aspect relation modeling and a label drop mechanism. LDNet assigns different relations to different levels for understanding and decision-making, thereby reducing decision confusion. By introducing the label drop mechanism, LDNet alleviates the influence of irrevelant relations.
Experimental results show that LDNet achieves highly competitive performance across 9 tasks, in both single-modal and multi-modal, few-shot and zero-shot settings, which verifies its effetiveness and universality.

\section*{Limitations}
 1) The total quantity and variety of MIE datasets are not enough, so LDNet cannot be pre-trained on a relatively large-scale dataset for MIE as it can be for IE, and since LDNet is a universal IE and UIE solution, its performance on certain multi-modal datasets may not be as good as models specifically designed for multi-modal tasks. 2) Due to the maximum input length constraint, LDNet may experience a performance decline in document-level information extraction.

\section*{Ethical Considerations}\label{societal-impacts}
If the model is able to extract information of high quality, it may be able to extract personal privacy information such as names, addresses, and phone numbers from large text datasets. This information could potentially be used for illegal monitoring, harassment, and other malicious purposes.
Establishing appropriate privacy protection mechanisms and usage restrictions can be applied to ensure that the extracted information is only used for legitimate purposes and not abused.

\bibliography{acl_latex}

\appendix
\section{Methodology}
\subsection{Schema Labels}\label{appendix-schema-labels}




When it comes to the specific implementation, the schema labels are divided into two parts: a set of special tokens \texttt{[LM], [LR], [LC], [TL], [TP],} and \texttt{[B]} and their corresponding label text. 

\texttt{[LM], [LR],} and \texttt{[LC]} are the special tokens that represent labels of entity mention, relation, and text classification respectively. Only one of the three special tokens \texttt{[TL], [TP],} and \texttt{[B]} will appear in a single data instance. \texttt{[TP]} is used for the MRC task, \texttt{[B]} is used for the classification task, and \texttt{[TL]} is used for all other tasks. 

The special tokens \texttt{[LM], [LR], [LC]} will be followed by their corresponding label text, which are the tokenized strings for the entity mention, relation, and text classification labels. All the labels that appear in the dataset will be included in the schema labels, such as \texttt{[LC], ``\_correct'', [LC], ``\_wrong''} for the classification task with ``correct'' and ``wrong'' as the two labels. 

We use the special tokens in the schema labels as the trigger words and utilize them to guide the relation extraction process. A relation will only be extracted when the trigger word is activated. The label drop operation sets the elements in the label vector corresponding to the special tokens that in gold spans to 1, while the other schema label elements remain 0. For example, in the classification task, if a data instance is classified as ``correct'', the element corresponding to the \texttt{[LC]} token before ``\_correct'' token will be set to 1, while the element corresponding to the ``\_correct'' token will remain 0.

\subsection{Handling of Unknown Schemas}\label{appendix-unknown-schemas}

If the schema is unknown, LDNet removes the schema from the original NER input format, transforming it from the format of \textit{instruction + schema + text}:

\begin{verbatim}
[I] Please identify possible entities 
from the given text and determine 
their types [LM] person [LM] location
[LM] organization [TL] Jerry Smith 
is a friend of Tom
\end{verbatim}

to:

\begin{verbatim}
[I] Please identify possible entities 
from the given text and determine 
their types [TP] Jerry Smith 
is a friend of Tom
\end{verbatim}

Here, \texttt{[I] Please identify possible entities from the given text and determine their types} is the instruction part, with \texttt{[I]} being a special token indicating the start of the instruction; \texttt{[LM] person [LM] location [LM] organization} represents the schema, with \texttt{[LM]} being a special token representing an entity type, such as \texttt{[LM] person}, indicating the person type; \texttt{[TL] Jerry Smith is a friend of Tom} is the text part, with \texttt{[TL]} being a special token indicating that the text following it requires not only span extraction but also the extraction of types within the schema. For instance, to extract the entity \texttt{Jerry Smith}, it's necessary to extract both its span and its corresponding type, \textbf{person}, which is associated with the span of the \texttt{[LM]} token in \texttt{[LM] person} in the schema. In the input without a schema, \texttt{[TP]} indicates that only entity spans need to be extracted from the text, without requiring the extraction of additional information such as types in schemas.

Thus, LDNet handles unknown schemas by processing the input as follows:

\begin{verbatim}
[I] Please identify possible entities 
from the given text and determine their 
types 
[TP] Jerry Smith is a friend of Tom
\end{verbatim}

\begin{table*}[h]
\small
    \centering
    \resizebox{0.6\textwidth}{!}{
    \begin{tabular}{cccc}
        \toprule
        Parameter & Value & Parameter & Value  \\
        \midrule
            warmup proportion & 0.1 & batch size & 4\\
            pre-training epochs & 3 & PLM learning rate & 2e-5 \\
            fine-tuning epochs & 20 &PLM weight decay & 0.1\\
            fine-tuning epoch patience & 3 &others learning rate & 1e-4\\
            few-shot epochs & 200 &max gradient norm & 1.0\\
            few-shot epoch patience & 10 &$d_h$ & 1024\\
            $d_v$$^{\diamondsuit}$ & 1024 & batch size$^{\diamondsuit}$ & 32 \\
            PLM learning rate$^{\diamondsuit}$ & 3e-5 & others learning rate$^{\diamondsuit}$ & 1e-4 \\
            fine-tuning epochs$^{\diamondsuit}$ & 20 & PLM weight decay$^{\diamondsuit}$ & 0.1 \\
            warmup proportion$^{\diamondsuit}$ & 0.01 & $\alpha^{\diamondsuit}$ & 0.5 \\
                \bottomrule
    \end{tabular}}
    \caption{The parameters marked with $^{\diamondsuit}$ are for the multi-modal experiments.}
    \label{params}
\end{table*}
\begin{table*}[]
    \centering
    \resizebox{0.9\textwidth}{!}{
    \begin{tabular}{cccccc}
        \toprule
        NER Pre-training Dataset & Instruction & Instance & RE Pre-training Dataset & Instruction & Instance \\
        \midrule
        AnatEM & 42 & 5,861 &  ADE\_corpus &  9 & 3,417 \\
        bc2gm & 42 & 12,500 & FewRel &  9 & 20,000 \\
        bc4chemd & 42 & 20,000 & GIDS &  9 & 8,526 \\
        bc5cdr & 42 & 4,560 & kbp37 &  9 & 15,807 \\
        Broad\_Tweet\_Corpus & 42 & 5,334 & New-York-Times-RE &  9 & 20,000 \\
        FabNER & 42 & 9,435 & NYT11HRL &  9 & 20,000 \\
        FindVehicle & 42 & 20,000 & semeval &  9 & 8,000 \\
        GENIA & 42 & 15,023 & WebNLG &  9 & 5,019 \\
        HarveyNER & 42 & 3,967 & Wiki-ZSL &  9 & 23,107 \\
        MultiNERD & 42 & 20,000 & MNRE$^{\diamondsuit}$ &  9 & 12,248 \\
        \cmidrule{4-6}
        NCBIdiease & 42 & 5,432 & MRC Pre-training Dataset & Instruction & Instance \\
        \cmidrule{4-6}
        ontoNotes5 & 42 & 20,000 &  BiPaR & 11,524 & 11,668 \\
        TweetNER7 & 42 & 7,103 & ms\_marco\_v2.1 & 20,000 & 20,000 \\
        WikiANN\_en & 42 & 20,000 & newsqa & 19,659 & 20,000 \\
        WNUT-16 & 42 & 2,394 & squad\_v2 & 19,998 & 20,000 \\
        Twitter2015$^{\diamondsuit}$ & 42 & 4,000 & SubjQA & 4,060 & 13,990 \\
        \cmidrule{4-6}
        Twitter2017$^{\diamondsuit}$ & 42 & 3,376 & EE Pre-training Dataset & Instruction & Instance \\
        \cmidrule{4-6}
        & & & PHEE & 40 & 2,898 \\
        
            \bottomrule
    \end{tabular}
    }
    \caption{The datasets marked with $^{\diamondsuit}$ are for the multi-modal experiments.}
    \label{ner-pretrain-datasets}
\end{table*}
\begin{table*}
    \centering
    \resizebox{0.98\textwidth}{!}{
    \begin{tabular}{cccccc}
        \toprule
    Classification Pre-training Dataset & Instruction & Instance & Classification Pre-training Dataset & Instruction & Instance \\
        \midrule
        ag\_news & 5 & 5,000 & ANLI  & 29 & 15,000\\
        ARC & 3,361 & 3,370 & CoLA & 43 & 5,000 \\
        CosmosQA & 4,483 & 5,000 & cos\_e & 5,000 & 5,000 \\
        dbpedia & 6 & 5,000 & DREAM & 3,842 & 5,000 \\
        hellaswag & 20 & 5,000 & IMDB & 26 & 5,000 \\
        MedQA & 5,000 & 5,000 & MNLI & 29 & 5,000 \\
        MRPC & 40 & 3,668 & MultiRC & 4,999 & 5,000 \\
        OpenBookQA & 4,835 & 4,957 & QASC & 4,832 & 5,000 \\
        QNLI & 31 & 5,000 & QQP & 40 & 5,000 \\
        RACE & 4,482 & 5,000 & RACE-C & 4,782 & 5,000 \\
        ReClor & 3,368 & 4,638 & RTE & 29 & 2,490 \\
        SciQ & 4,989 & 5,000 & SNLI & 29 & 5,000 \\
        SST-2 & 26 & 5,000 & Winogrande & 20 & 5,000 \\   
        WNLI & 31 & 635 \\
            \bottomrule
    \end{tabular}
    }
    \caption{The Classification Pre-training Datasets.}
    \label{classification-pretrain-datasets}
\end{table*}
\begin{table*}[h]
\centering
\resizebox{\textwidth}{!}{
\begin{tabular}{c|ccccccc}
\toprule
Model & TANL & UIE & DeepStruct & InstructUIE & USM & Mirror & LDNet \\
\midrule
\multicolumn{1}{c|}{\multirow{1}{*}{Computational Complexity}} & $ O(n^3) $ & $ O(n^2) $ & $ O(n^2) $ & $ O(n^2) $ & $ O(n^2) $ & $ O(n^2) $ & $ O(n^2) $ \\
\multicolumn{1}{c|}{\multirow{1}{*}{PLM}} & T5-base & T5-large & GLM & FlanT5 & RoBERTa-large & DeBERTa-v3-large & DeBERTa-v3-large \\
\multicolumn{1}{c|}{\multirow{1}{*}{PLM Params}} & 220M & 770M & 10B & 11B & 372M & 304M & 304M \\
\bottomrule
\end{tabular}%
}
\caption{Computational complexity and model parameters of various models.}
\label{model-complexity}
\end{table*}
\begin{table*}[h]
  \centering
  \resizebox{\textwidth}{!}{%
  \begin{tabular}{c|c|cccccc|cc}
    \toprule
    Task & Datasets & TANL & UIE  & USM & FSUIE-base & UniEX-large & MetaRetriever & LDNet$_{deberta-v3-base}$
      & LDNet$_{deberta-v3-large}$ \\
    \midrule
    \multirow{3}{*}{NER}
     & ACE04   & -     & 86.89  & 87.62  & 85.24 & 87.12 & 86.10   & 88.50 &  88.69 \\
     & ACE05   & 84.90 & 85.78  & 87.14  & 86.22 & 87.02 & 84.01   & 88.18 & 87.79 \\
     & CoNLL03 & 91.70 & 92.99  & 93.16  & - & 92.65& 92.38  & 93.43 & 93.44 \\ 
      \midrule
      \multirow{4}{*}{RE}
     & ACE05   & 63.70 & 66.06 & 67.88  & 72.29 & 66.06 &64.37    & 69.33 & 69.14 \\
     & CoNLL04 & 71.40 & 75.00 & 78.84   & -& 73.40 & 73.66   & 79.26 & 80.79 \\
     & NYT     & -     & 93.54  & 94.07   &- &- & -   & 94.15 & 94.96 \\
     & SciERC  & -     & 36.53  & 37.36  & - & 38.00 &35.77   & 37.49 & 46.85 \\
      \midrule
      \multirow{4}{*}{EE} &
      ACE05-Tgg  & 68.40 & 73.36  & 72.41  &- &74.08 &72.38   & 73.42 & 83.56\\
     & ACE05-Arg & 47.60 & 54.79  & 55.83  &- &53.92 & 52.62   & 57.04  &61.14\\
     & CASIE-Tgg & -     & 69.33 & 71.73    &- &71.46 &69.76    & 72.77 & 73.36 \\
     & CASIE-Arg & -     & 61.30  & 63.26    &- &62.91 & 60.37   & 63.35 & 63.88 \\
      \midrule
      \multirow{4}{*}{ABSA}
     & 14-res & - & 74.52 & 77.26   & 74.17 & 74.77& 73.41    & 79.08 & 79.17 \\
     & 14-lap & - & 63.88  & 65.51  & 65.56 & 65.23 & 62.83    & 68.84 & 69.00 \\
     & 15-res & - & 67.15  & 69.86   & 70.63 &68.58 & 65.85   & 74.47 & 71.18 \\
     & 16-res & - & 75.07  & 78.25   & 75.80 & 76.02 & 73.55    & 78.40 &78.85\\
    \bottomrule
    \end{tabular}%
  }
  \caption{Comparison to the performance of models with smaller pre-trained language models.}
  \label{smaller-plms}
\end{table*}
\begin{table*}[]
  \centering
  \resizebox{\textwidth}{!}{%
  \begin{tabular}{c|c|cccc|ccc}
    \toprule
    Task & Datasets &GoLLIE Baseline & GoLLIE & GoLLIE-13B & GoLLIE-34B & LDNet Baseline& LDNet$_{deberta-v3-base}$
      & LDNet$_{deberta-v3-large}$ \\
    \midrule
    \multirow{2}{*}{NER}
     & ACE05   & 89.10 & 88.10 & 89.40 & \textbf{89.60} & 85.70 & 88.18 & 87.79  \\
     & CoNLL03 & 92.90 & 92.80 & 93.00 & 93.10 & 92.73 & 93.43 & \textbf{93.44} \\ 
      \midrule
      \multirow{1}{*}{RE}
     & ACE05   & 63.80 & 63.60 & 67.50 & \textbf{70.10} & 67.86 & 69.33 & 69.14 \\
      \midrule
      \multirow{4}{*}{EE} &
      ACE05-Tgg  &71.7 & 72.2 & 70.9 & 71.9 & 73.05 & 73.42 & \textbf{83.56}\\
     & ACE05-Arg &65.9 & 66.0 & 67.8 & \textbf{68.6} &54.73 & 57.04  &61.14\\
     & CASIE-Tgg & 33.9 & 59.3 & 62.2 & 65.5 &71.60 & 72.77 & \textbf{73.36} \\
     & CASIE-Arg & 47.9 & 50.0 & 52.6 & 55.2 &61.04 & 63.35 & \textbf{63.88} \\
    \bottomrule
    \end{tabular}%
  }
  \caption{Comparison to GoLLIE models.}
  \label{tab:gollie-results}
\end{table*}
From the text \textit{"Jerry Smith is a friend of Tom"}, LDNet directly extracts the spans of entities such as \texttt{Jerry Smith} and \texttt{Tom} as the output.

\subsection{Handling of Discontinuous NER and Nested NER}\label{appendix-discontinuousNER}
\textbf{Discontinuous NER} LDNet handles discontinuous NER in a manner similar to how it handles regular NER. Suppose the input is:

\begin{verbatim}
[I] Please identify possible entities 
from the given text 
and determine their types 
[LM] person [LM] title 
[LM] organization [TL] The CEO of Tesla
, Elon Musk, made an announcement today
\end{verbatim}

LDNet will extract the following relations:
\begin{itemize}
    \item TA Relation: Between \textit{\texttt{[LM]}} before \textbf{person} and \textit{\texttt{CEO of Tesla}}, and \textit{\texttt{[LM]}} before \textbf{person} and \textit{\texttt{Elon Musk}}.
    \item A2A Relation: Between \textit{\texttt{CEO of Tesla}} and \textit{\texttt{Elon Musk}}.
    \item AS Relation: Between \textit{\texttt{CEO of Tesla}} and \textit{\texttt{CEO of Tesla}}, and \textit{\texttt{Elon Musk}} and \textit{\texttt{Elon Musk}}.
\end{itemize}

The closed loop formed by the TA relation between \textit{\texttt{[LM]}} before \textbf{person} and \textit{\texttt{CEO of Tesla}}, the AS relation between \textit{\texttt{CEO of Tesla}} and \textit{\texttt{CEO of Tesla}}, the A2A relation between \textit{\texttt{CEO of Tesla}} and \textit{\texttt{Elon Musk}}, the AS relation between \textit{\texttt{Elon Musk}} and \textit{\texttt{Elon Musk}}, and the TA relation between \textit{\texttt{[LM]}} before \textbf{person} and \textit{\texttt{Elon Musk}} allows LDNet to extract \texttt{CEO of Tesla, Elon Musk} as a discontinuous entity of the \textbf{person} type.

\textbf{Nested NER} The input format for nested NER is:

\begin{verbatim}
[I] Please identify possible entities 
from the given text and 
determine their types
[LM] organization [LM] title 
[LM] person [LM] location 
[TL] Apple CEO Tim Cook 
gave a speech at Stanford University 
in California
\end{verbatim}

In this example, \texttt{Apple} is an organization entity representing the Apple company, and \texttt{Apple CEO} is a title entity representing the CEO position of Apple.

After obtaining the TA, A2A, and AS relation probability matrices, LDNet will extract the following relations:

\begin{itemize}
    \item TA Relation: Between \textit{\texttt{[LM]}} before \textbf{organization} and \textit{\texttt{Apple}}, between \textit{\texttt{[LM]}} before \textbf{title} and \textit{\texttt{Apple}}, and \textit{\texttt{[LM]}} before \textbf{title} and \textit{\texttt{CEO}}.
    \item AS Relation: Between \textit{\texttt{Apple}} and \textit{\texttt{Apple}}; between \textit{\texttt{Apple}} and \textit{\texttt{CEO}}.
\end{itemize}

The closed loop formed by the TA relation between \textit{\texttt{[LM]}} before \textbf{organization} and \textit{\texttt{Apple}}, and the AS relation between \textit{\texttt{Apple}} and \textit{\texttt{Apple}} allows LDNet to extract \textit{\texttt{Apple}} as an organization entity. Similarly, the TA relations between \textit{\texttt{[LM]}} before \textbf{title} and \textit{\texttt{Apple}}, the AS relation between \textit{\texttt{Apple}} and \textit{\texttt{CEO}}, and the TA relation between \textit{\texttt{[LM]}} before \textbf{title} and \textit{\texttt{CEO}} form a closed loop, allowing LDNet to extract \textit{\texttt{Apple CEO}} as a title entity.

The A2A relation is not mandatory, as the information to be extracted may not involve two arguments, such as in NER tasks.



\subsection{Underlying Logic of Label Drop Mechanism}\label{label-drop-logic}
The underlying logic of label drop mechanism is that if the $i$-$th$ token does not exist in the gold answer, after training, the value of $\hat{l}^r_i$ at position $i$ in $\hat{l}^r$ will close to 0, it suppresses the value of $s_{.i}^r$ in the $i$-th column of the $S^{r}$ matrix.
During subsequent decoding, if the value of $s_{.i}^r$ is suppressed below a threshold, LDNet will not consider extracting the relation between position $i$ and other positions, therefore filtering out token pairs that is impossible to have relations. 
As shown in Figure~\ref{fig:model-framework}, 
for the convenience of observation, we only depict the prediction of the A2A relation. And to better illustrate the concept of label drop, we hypothesize an extreme scenario where $\hat{l}^r$ is the same as $\mathbf{l}^r$. For instance, the probability of a relation existing between ``playwright'' and ``born'' is set to 0.

\subsection{LDNet's Contributions Relative to Prior Multi-modal Approaches}
Multi-modal representation is a conventional approach. LDNet utilizes multi-modal representation to enable its application to multi-modal data. LDNet is a universal information extraction method, its capability is general and not differentiated by whether the data is multi-modal or unimodal. Compared to prior multi-modal approaches, the contribution of LDNet lies not in combining multi-modal information, but rather in its Label Drop mechanism, which effectively filters out irrelevant token pairs. 

\paragraph{Individual Contributions of Image Features} Since images only provide clue information; the information extraction ultimately comes from the text. Therefore, it is not possible to explore the individual contributions of image features on the model's performance.

\section{Related Work}
\paragraph{Dropout Strategy}
Dropout~\citep{dropout} is a powerful technique usually used to regularize the training of deep neural networks. R-Drop~\citep{rdrop} forces the output distributions of different submodels, sampled by dropout, to be consistent with each other by minimizing their bidirectional KL-divergence.
LDNet transfers the idea of dropout into IE tasks and applies label drop to remove unrelational token pairs, forcing the model to concentrate on relational ones.

\begin{table*}[]
  \centering
  \small
  \resizebox{0.75\textwidth}{!}{
  \begin{tabular}{l|c|ccccccc}
    \toprule
    \multirow{2}{*}{Model} & SQuAD 2.0 & CoLA & QQP & MNLI & SST-2 & QNLI & RTE & MRPC \\
    & (EM/F1) & (Mcc) & (Acc) & (Acc) & (Acc) & (Acc) & (Acc) & (Acc) \\
    \midrule
    BERT-large         & 79.0 / 81.8         & 60.6       & 91.3      & -          & 93.2        & 92.3       & 70.4      & 84.1       \\
    RoBERTa-large      & 86.5 / 89.4         & 68.0       & 92.2      & 90.2       & 96.4        & 93.9       & 86.6      & 88.8       \\
    DeBERTa v3-large   & 89.0 / 91.5         & 75.3       & 93.0      & 91.9       & 96.9        & 96.0       & 92.7      & 92.2       \\
    Mirror & 40.4 / 67.4         & 63.9       & 84.8      & 85.9       & 93.6        & 91.6       & 85.9      & 89.2 \\
    LDNet & 42.5 / 72.0 & 74.2& 86.2 &87.2 &94.8 &92.8 &89.2 & 91.0\\
    \bottomrule
  \end{tabular}}
  \caption{
    Results on MRC and classification tasks.
  }
  \label{tab:mrc-cls-results}
\end{table*}
\paragraph{Model Transfer Learning}

Averaging the predictions of all trained models is a simple yet effective way to enhance the performance of almost any machine learning algorithm. However, it can be cumbersome and computationally expensive. Therefore, \cite{kd} proposed Knowledge Distillation (KD) to compress the knowledge of an ensemble into a single model. 
\cite{docredkd} utilizes KD in document-level relation extraction. The system trains a teacher model on the distantly-supervised data and uses the distributions generated by the teacher model as soft labels to pre-train the student model. The authors found that distilling with the MSE loss performs better than using KL-divergence. LDNet incorporates this idea and follows the same setting, using MSE loss to minimize the difference between the distributions of the teacher model and the student model.
\begin{table*}[h]
    \centering
    \renewcommand{\arraystretch}{1.5}
    \resizebox{0.85\textwidth}{!}{%
    \begin{tabular}{c|c|cccccc|ccc}
        \toprule
        \multirow{2}{*}{\textbf{Modality}} & \multirow{2}{*}{\textbf{Methods}} & \multicolumn{3}{c}{Twitter-2015} & \multicolumn{3}{c|}{Twitter-2017} & \multicolumn{3}{c}{MNRE} \\
        \cmidrule{3-11}
         &  & Prec. & Rec. & F1 & Prec. & Rec. & F1 & Prec. & Rec. & F1 \\
        \midrule
         
        \multirow{10}{*}{Text+Image} &
        AdapCoAtt~\citep{AdapCoAtt} & 69.87 & 74.59 & 72.15 & 85.13 & 83.20 & 84.10 & - & - & - \\
        &VisualBERT~\citep{visualbert}& 68.84 &71.39& 70.09 &84.06 &85.39 &84.72 &57.15& 59.48 &58.30\\
         & OCSGA~\citep{ocsga} & 74.71 & 71.21 & 72.92 & - & - & - & - & - & - \\
         & UMT~\citep{umt} & 71.67 & 75.23 & 73.41 & 85.28 & 85.34 & 85.31 & 62.93 &63.88 & 63.46 \\
         & UMGF~\citep{umgf} & 74.49 & 75.21 & 74.85 & 86.54 & 84.50 & 85.51 & 64.38 & 66.23 & 65.29\\
         & MEGA~\citep{mnre} & 70.35 & 74.58 & 72.35 & 84.03 & 84.75 & 84.39 & 64.51 & 68.44 & 66.41 \\
         & HVPNeT~\citep{hvpnet} & 73.87 & 76.82 & 75.32 & 85.84 & 87.93 & 86.87 & 83.64 &80.78& 81.85\\
         & MoRe~\citep{more} & - & - & - & - & - & - & 65.25 & 67.32 & 66.27\\
         & TMR~\citep{tmr} & 75.26 & 76.49 & 75.87 & 88.12 & 88.38 & 88.25 & 90.48 & 87.66 & 89.05\\ \cmidrule{2-11}
         & LDNet & 76.79&75.56&76.17&87.51&87.64&87.57& 75.45  &76.15 & 75.79\\   
        \midrule
    \end{tabular}%
    }
    \caption{Comparison of LDNet's performance on MIE tasks with some MIE baselines.}
    \label{multi-appendix}
\end{table*}
\begin{table*}[t]
\centering
\resizebox{0.85\textwidth}{!}{
\begin{tabular}{c|ccccccccccc}
\toprule
Dataset & Strategy & 10\% & 20\% & 30\% & 40\% & 50\% & 60\% & 70\% & 80\% & 90\% & 100\% \\
\midrule
\multicolumn{1}{c|}{\multirow{2}{*}{CoNLL03}} & w / PT \& Inst. & 94.74 & 94.35 & 94.29 & 94.49 & 95.04 & 94.68 & 94.08 & 94.56 &94.29 & 94.60 \\
\multicolumn{1}{l|}{}  &w / PT \& w/o Inst.& 94.69 & 94.11 & 95.06 & 93.49 & 94.57 & 94.62 & 94.92 & 94.65 & 94.88 & 95.11 \\
\midrule
\multicolumn{1}{c|}{\multirow{2}{*}{NYT}}& w / PT \& Inst.  & 93.72 & 93.51 & 93.36 & 93.16 & 93.59 & 93.07 & 93.38 & 93.48 & 92.86 & 94.43 \\
\multicolumn{1}{l|}{}  &w / PT \& w/o Inst. & 93.35 & 93.86 & 93.17 & 93.39 & 93.96 & 93.70 & 93.54 & 93.25 & 93.54 & 94.41 \\\midrule
\multicolumn{1}{c|}{\multirow{2}{*}{ACE05-Tgg}}& w / PT \& Inst.  & 71.79 & 72.59 & 68.33 & 74.45 & 67.23 & 70.07 & 69.14 & 75.58 & 68.66 & 78.74 \\
\multicolumn{1}{l|}{}  &w / PT \& w/o Inst. & 67.10 & 66.67 & 69.57 & 68.61 & 76.43 & 79.55 & 70.92 & 71.25 & 78.61 & 80.23 \\\midrule
\multicolumn{1}{c|}{\multirow{2}{*}{ACE05-Arg}}& w / PT \& Inst. & 49.80 & 48.73 & 60.42 & 50.24 & 58.95 & 47.06 & 49.64 & 51.22 & 42.49 & 55.78 \\
\multicolumn{1}{l|}{}  &w / PT \& w/o Inst.& 50.95 & 46.15 & 42.93 & 52.22 & 50.00 & 60.47 & 48.70 & 54.94& 55.77 & 62.76 \\\midrule
\multicolumn{1}{c|}{\multirow{2}{*}{14-res}}& w / PT \& Inst. & 84.62 & 83.86 & 84.29 & 85.58 & 84.82 & 83.93 & 85.37 & 84.75 & 85.10 & 83.58 \\
\multicolumn{1}{l|}{}  &w / PT \& w/o Inst.& 84.54 & 82.97 & 84.21 & 84.88 & 84.60 & 84.82 & 84.49 & 84.13 & 82.66 & 84.79 \\
\bottomrule
\end{tabular}}
\caption{Results of the label drop mechanism with different drop rates.}
\label{tab:label_drop_results}
\end{table*}
\begin{table*}[]
\centering
\resizebox{0.8\textwidth}{!}{
\begin{tabular}{c|ccccccccccc}
\toprule
Dataset & Metric & 10\% & 20\% & 30\% & 40\% & 50\% & 60\% & 70\% & 80\% & 90\% & 100\% \\
\midrule
\multicolumn{1}{c|}{\multirow{3}{*}{Twitter-2015}} & P & 76.22 &76.52 & 78.22 & 77.61 &76.36 &75.93 &75.56 & 79.45& 76.74& 76.79\\
\multicolumn{1}{l|}{} & R & 76.32 &75.60  & 74.36 & 73.61 &76.19 & 75.06 & 75.86 & 74.03& 76.21& 75.56\\
\multicolumn{1}{l|}{} & F1 & 76.27 & 76.05 & 76.24 & 75.56 & 76.28& 75.49 &75.71  & 76.64& 76.47 & 76.17\\\midrule
\multicolumn{1}{c|}{\multirow{3}{*}{Twitter-2017}} & P & 87.80 & 87.33 & 89.10 & 88.25 & 88.7 & 87.69&86.39	&87.33 & 86.49&87.51\\
\multicolumn{1}{l|}{} & R &87.34 &86.75 & 86.68 & 86.75 & 87.27&87.49	&87.86&87.79 & 87.19&87.64 \\
\multicolumn{1}{l|}{} & F1 & 87.57 & 87.04 & 87.88 & 87.50 & 87.99&87.59&87.12&87.56&86.84& 87.57\\ \midrule
\multicolumn{1}{c|}{\multirow{3}{*}{MNRE}} & P & 74.21 &75.16 & 76.05 &75.29 & 75.80&74.74&74.73&74.54&73.57 & 75.45\\
\multicolumn{1}{l|}{} & R & 73.79 & 74.97 & 72.99 & 72.49 &  74.72&74.41&74.04&72.37& 72.61& 76.15\\
\multicolumn{1}{l|}{} & F1 & 74.00 & 75.06 & 74.49 &73.86 &  75.26&74.57	&74.39&73.44& 73.09& 75.79\\ 
\bottomrule
\end{tabular}%
}
\caption{Results of the label drop mechanism with different drop rates on MIE datasets.}
\label{tab:multi-modal-label_drop_results}
\end{table*}
\begin{figure*}
\begin{minipage}{0.28\textwidth}
\centering
\resizebox{\textwidth}{!}{
\begin{tabular}{cl|c|c}
\toprule
\multicolumn{2}{c|}{w/ Label Drop} & \textcolor{red}{\xmark}  & \color[HTML]{008114}\cmark  \\ 
\midrule
\multicolumn{1}{c|}{\multirow{3}{*}{Twitter-2015}}   & P     & 74.07 & 76.79   \\
 \multicolumn{1}{l|}{}  &R  &60.57 &75.56 \\ 
  \multicolumn{1}{l|}{}  &F1  &66.64 &76.17\\ 
 \midrule
\multicolumn{1}{c|}{\multirow{3}{*}{Twitter-2017}}   & P     & 84.42 &87.51   \\
 \multicolumn{1}{l|}{}  &R  &81.54 &87.64\\ 
  \multicolumn{1}{l|}{}  &F1   & 82.96&87.57 \\ 
 \midrule
 \multicolumn{1}{c|}{\multirow{3}{*}{MNRE}}   & P     & 74.04 &75.45   \\
 \multicolumn{1}{l|}{}  &R  & 68.66 & 76.15\\ 
  \multicolumn{1}{l|}{}  &F1 & 71.25 &75.79\\ 
\bottomrule
\end{tabular} 
}
\end{minipage}
\begin{minipage}{0.33\textwidth}
  \includegraphics[width=\textwidth]{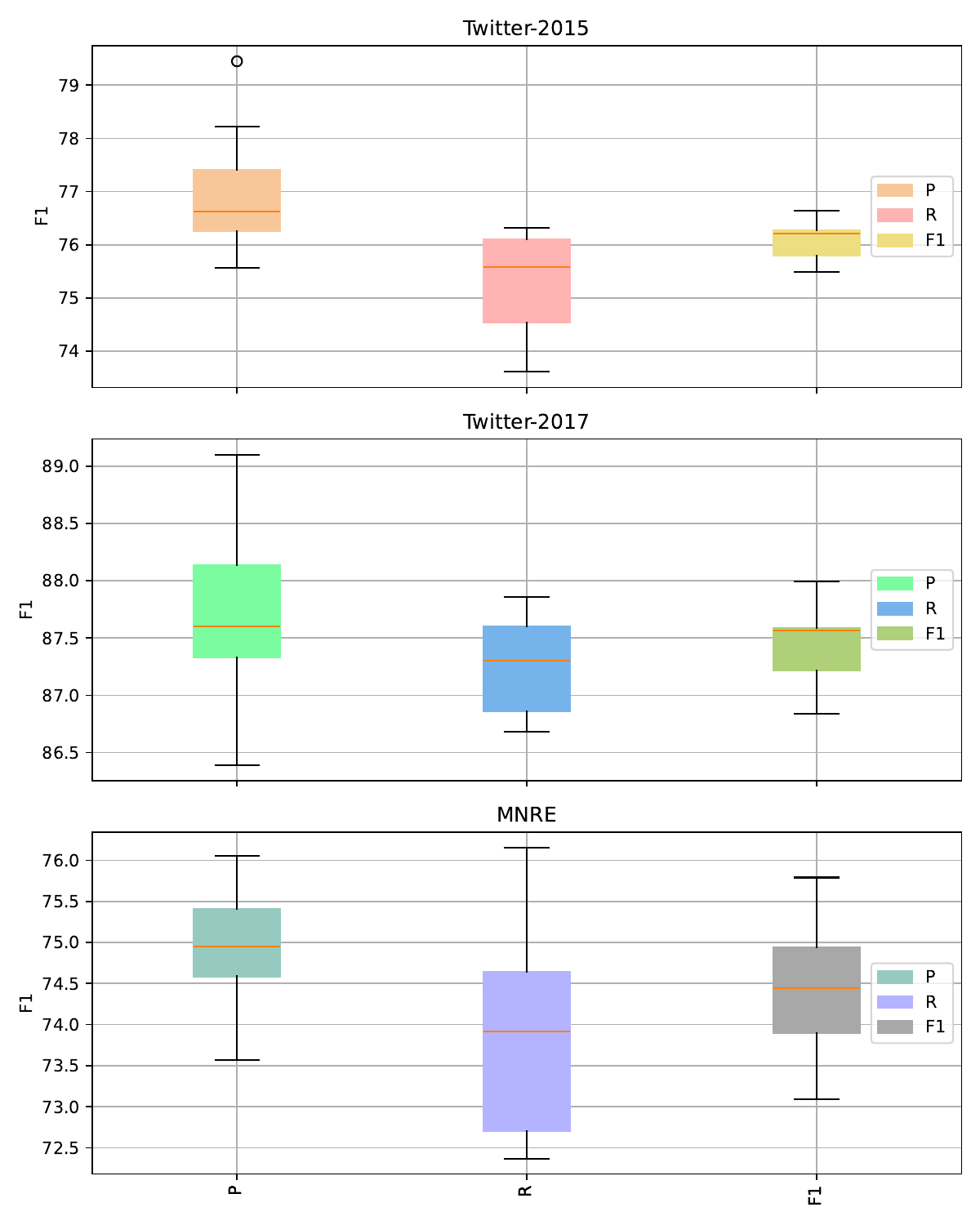}
\end{minipage}
\begin{minipage}{0.35\textwidth}
  \includegraphics[width=\textwidth]{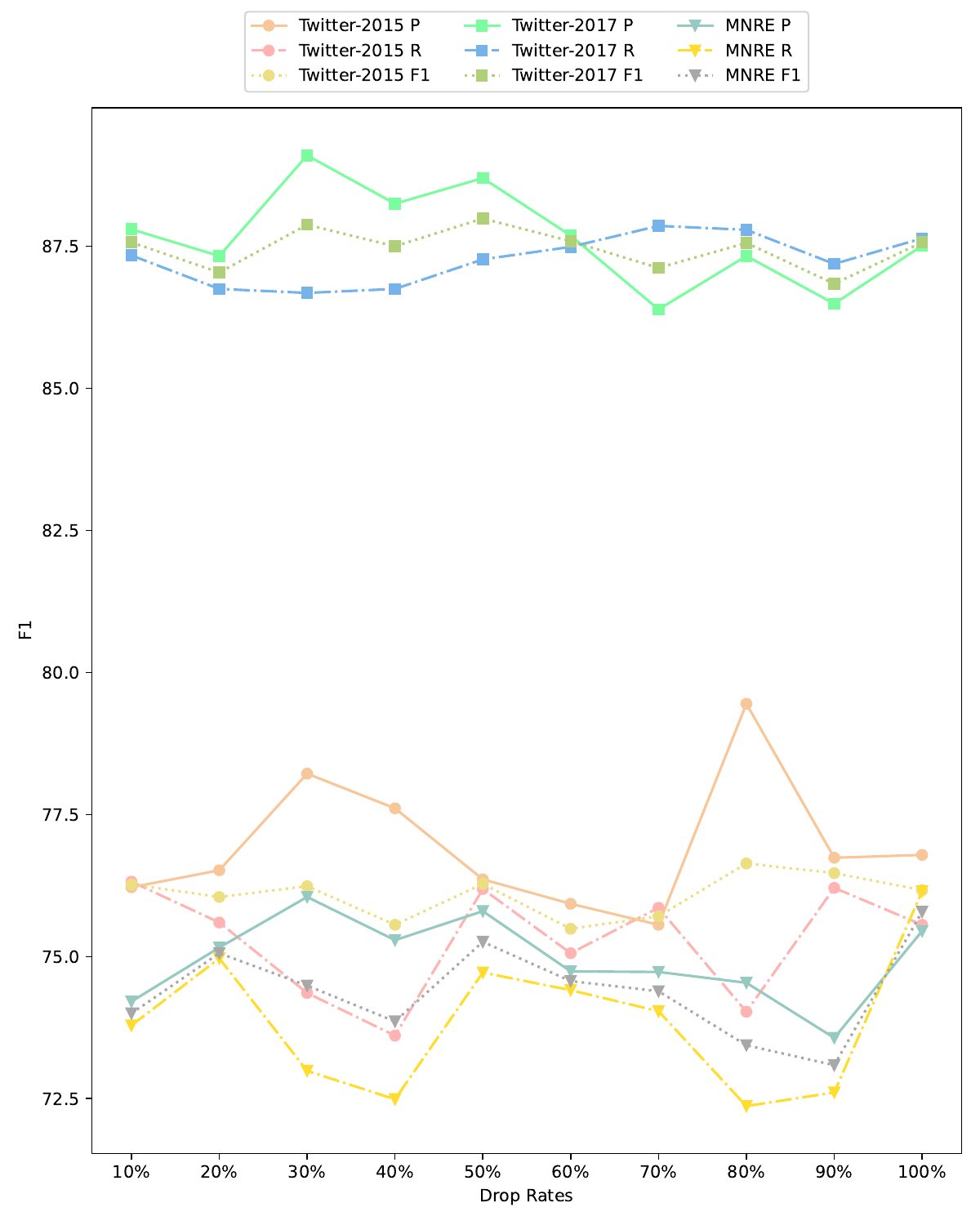}
\end{minipage}
    \caption{
        Results of the ablation study on the label drop mechanism on MIE tasks. 
    }
    \label{fig:image_and_table2}
\end{figure*}

\paragraph{Multi-modal Information Extraction} Unlike traditional information extraction, which relies exclusively on single-modal data, Multi-modal Information Extraction (MIE) leverages auxiliary visual cues from images to supplement the missing context. MEGA~\citep{mnre} is the first to propose the MNRE dataset and introduces a dual image alignment method to capture aligned information between visual object and textual object. 
UMT~\citep{umt}, which pioneers the use of Transformer in the MIE task, utilizes a multi-modal interaction module to integrate token representations with visual representations. 
MoRe~\citep{more} enhances textual information retrieval by leveraging images and titles from search engines, thereby improving the accuracy of multimodal RE and NER tasks. Building on MoRe, MRE-RS~\citep{mrers} retrieves textual and visual evidence at different levels and further proposes a novel method to synthesize information for improved reasoning across the same and different modalities. CoTPD~\citep{cotpd} demonstrates the elicitation of reasoning abilities from LLMs using CoT prompts across various dimensions and introduces a conditional prompt distillation method to transfer commonsense reasoning to a student model, enhancing its performance on text-only inputs. There are also some recently new methods, such as UMIE~\citep{umie} and  OmniParser~\citep{omniparser}.

\section{Experiments details}
\subsection{Hyperparameters and Pre-training Datasets}\label{hyperparams}
Specifically, we use vit-large-patch32-224-in21k as our image backbone. The experiments can be run using only 1 NVIDIA RTX 3090 with 24 GB memory. The hyperparameters and pre-training datasets are in Table~\ref{params}, Table~\ref{ner-pretrain-datasets} and Table~\ref{classification-pretrain-datasets}.

\subsection{Experimental Configurations}\label{appendix-configurations}

The LDNet configurations `w/ PT w/o Inst.' and `w/ PT w/ Inst.' are not considered variants of the model; the only difference lies in whether the instruction part of the model input is an empty string. In `w/ PT w/o Inst.' configuration, the instruction part is an empty string, while in `w/ PT w/ Inst.', the instruction part is not empty and is a string similar to `Please determine the two entities mentioned in the text and specify the nature of their relationship.' 

We here give reasons why performance is often better in `without PT and Inst.' configurations. The situation that configurations without PT and Inst. perform better is related to the dataset and the backbone model, DeBERTa-v3-large. For certain datasets like CoNLL04, 14-lap, and 15-res, better performance is observed under the `with PT' configuration. On the other hand, some datasets, such as ACE04 and ACE05, are inherently large, and fine-tuning on their training sets alone yields good results, and the absence of pre-training does not introduce interference from other data, leading to better performance without PT. As shown in Table~\ref{tab:compare-mirror}, the performance difference between `with and without Inst.' is generally small. The distinction between these two configurations lies in whether the instruction part of LDNet's input is an empty string, and any difference should be related to DeBERTa-v3-large's instruction-following capability.

\subsection{Method for Few-shot Experiments}
Regarding the few-shot experiments, we employed fine-tuning. Following the settings of Mirror to ensure the fairness of the comparison, we fine-tuned for several epochs on the training set of the few-shot dataset before assessing few-shot capabilities.

\subsection{Model Parameters and Computational Complexity}\label{model-parameters-and-computational-complexity}
All models' computational complexities and the scales of the pre-trained language models used are listed in the Table~\ref{model-complexity}.

Here, we define the following notations to analyze the computational complexity of the models:
\begin{itemize}
    \item Expected answer length: $k$
    \item Text length: $n$
    \item Instruction length: $l$
    \item Schema length: $s$
    \item Hidden dimension: $d$
    \item Biaffine dimension: $b$
    \item Number of transformer layers: $t$
\end{itemize}

First, the computation for the hidden states of a transformer architecture model (excluding the computation for predicting token logits) is $(24nd^2 + 4dn^2) \cdot t$. The computation for predicting token logits is $2dvn$.

TANL, Deepstruct, UIE, and InstructUIE, which are generative unified information extraction methods, mainly focus on the computation involved in token prediction.

\begin{itemize}
    \item \textbf{TANL} outputs a sequence that includes the answer to be extracted in the original input sequence, so its computation for sequence generation is $[(24nd^2 + 4dn^2) \cdot t + 2dvn] \cdot (n + k)$. Since the decoding algorithm used by TANL has a complexity of $O(n^2)$, its overall computational complexity is $O(n^3)$.
    \item \textbf{Deepstruct} outputs a sequence that is the answer sequence itself, so its computation is $[(24nd^2 + 4dn^2) \cdot t + 2dvn] \cdot k$, with a computational complexity of approximately $O(n^2)$.
    \item \textbf{UIE} takes an input sequence that includes the schema and directly generates a Structured Extraction Language (SEL) as output. Its computation is $[(24(n+s)d^2 + 4d(n+s)^2) \cdot t + 2dv(n+s)] \cdot k$, with a computational complexity of approximately $O(n^2)$.
    \item \textbf{InstructUIE} includes the instruction, schema, and text in the input sequence. Its computation is $[(24(n+s+l)d^2 + 4d(n+s+l)^2) \cdot t + 2dv(n+s+l)] \cdot k$, with a computational complexity of approximately $O(n^2)$.
\end{itemize}

USM, Mirror, and LDNet belong to extractive unified information extraction methods, hence they do not include the computation for predicting token logits.

\begin{itemize}
    \item \textbf{USM} includes schema and text in the input. In addition to encoder representations, it computes token-token linking scores, label-token linking scores, and token-label linking scores. Its computation is $(24(n+s)d^2 + 4d(n+s)^2) \cdot t + 8(n+s)d^2 + 2d(n+s)^2$, with a computational complexity of $O(n^2)$.
    \item \textbf{Mirror} includes instruction, schema, and text in the input. After computing the representations with the pre-trained model, it uses a biaffine transformation to generate score matrices. Its computation is $(24(n+s+l)d^2 + 4d(n+s+l)^2) \cdot t + 4ndb + 4nd + 6nb^2 + 6bn^2$, with a computational complexity of $O(n^2)$.
    \item \textbf{LDNet} includes instruction, schema, and text in the input. The Multi-aspect Relation Modeling module's computation is $(8nd^2 + 2dn^2) \cdot 3$. The label drop mechanism's computation is $3n^2$. Therefore, the total computation is $(24(n+s+l)d^2 + 4d(n+s+l)^2) \cdot t + (8nd^2 + 2dn^2) \cdot 3 + 3n^2$, with a computational complexity of $O(n^2)$.
\end{itemize}

\subsection{Additional Experiments}\label{multi-modal baselines}
\paragraph{Comparison to Models with Smaller PLMs} We separately list the pretrained language model parameters smaller than the DeBERTa-v3-Large (304M) used by LDNet and compare them with LDNet using DeBERTa-v3-Base (86M). The results are shown in the Table~\ref{smaller-plms}.

\paragraph{Comparison to GoLLIE Models} In the Table~\ref{tab:gollie-results}, we compare the results of GoLLIE and LDNet. For ease of comparison, we have included only the results where both models are evaluated. 

\paragraph{MRC and Classification Results}
To demonstrate the compatibility of LDNet, we conducted experiments on SQuAD v2~\citep{squad-v2} and the 7 GLUE datasets~\citep{cola,glue,mnli,sst-2,mrpc}. As shown in Table~\ref{tab:mrc-cls-results}, LDNet outperforms Mirror across all datasets. Additionally, LDNet surpasses BERT-large on SST-2 and QNLI, outperforms RoBERTa-large on CoLA, RTE, and MRPC, and achieves competitive results with DeBERTa v3-large on CoLA and MRPC.
It is important to note that LDNet is a universal solution for IE and does not undergo full fine-tuning like Language Model Models (LLMs). 
Therefore, LDNet has limitations when it comes to tasks such as MRC and classification. Consequently, it is reasonable to observe a slight performance gap in these tasks. However, it is worth noting that LDNet exhibits a smaller performance gap compared to Mirror.

\paragraph{Multi-modal Results} We put detailed results of multi-modal experiments in Table~\ref{multi-appendix}. It can be seen that LDNet performs better than most baselines, and is slightly lower in some metrics on certain specific baselines. But these methods are all focused only on MIE, unlike LDNet which is a universal solution for both IE and MIE. And some methods like HVPNeT utilizes additional visual prefixes and uses a specially-designed pyramid structure, and TMR have introduced more datasets for training, so there may be some performance gap. Although LDNet performs slightly lower than TMR on some datasets, it surpasses TMR on the Twitter-2015 dataset. Moreover, LDNet is a universal information extraction solution that covers a broader range of tasks, such as Discontinuous NER and Hyper RE.

\paragraph{Specific Results of Label Drop Mechanism}\label{drop-rate ablation}
We put the specific results of the label drop mechanism with different drop rates in Table~\ref{tab:label_drop_results}. It can be seen that after 100 rounds of fine-tuning, the performance of the two strategies on the CoNLL03, NYT, and 14-res datasets is not much different for LDNet, and in some drop rate cases, LDNet can perform better without the instruction, demonstrating the effectiveness of the label drop mechanism in the absence of specified instructions, which can be extended to other datasets without annotated instructions.

We also conduct ablation experiments on multi-modal datasets, and the experiments on multi-modal datasets are trained for 20 rounds, just like the main experiments. The results we release are the w/ PT \& Inst. results. From Table~\ref{tab:multi-modal-label_drop_results} and Figure~\ref{fig:image_and_table2}, we can see that the performance with the label drop mechanism is better than without it, which demonstrates the effectiveness of the label drop mechanism in MIE tasks. Under different drop rates, the F1 scores on the MIE datasets do not fluctuate greatly, indicating that the label drop mechanism still has robustness in MIE.

\end{document}